\documentclass{article}

\usepackage[nonatbib,final]{neurips_2020}

\usepackage[utf8]{inputenc} 
\usepackage[T1]{fontenc}    
\usepackage{hyperref}       
\usepackage{url}            
\usepackage{booktabs}       
\usepackage{amsfonts}       
\usepackage{nicefrac}       
\usepackage{microtype}      
\usepackage{float}
\usepackage{mathrsfs}
\usepackage{enumitem}
\usepackage{xcolor}
\usepackage{cite}
\usepackage{empheq}
\usepackage{textcomp}

\usepackage{amsthm, amssymb, amsmath}
\usepackage{verbatim, float}
\usepackage{mathrsfs}
\usepackage{graphics}
\usepackage{subfig}
\usepackage[usenames,dvipsnames]{pstricks}
\usepackage{multirow}
\usepackage{url}
\usepackage{hyperref}
\usepackage{array}
\usepackage{tabu}
\usepackage{epsfig}
\usepackage{parallel}
\usepackage{graphicx}
\usepackage{diagbox}
\usepackage{mathtools}
\usepackage{wrapfig}
\usepackage{makecell}
\usepackage[linesnumbered,lined,boxed,commentsnumbered]{algorithm2e}
\usepackage{diagbox}
\usepackage[absolute,overlay]{textpos}

\theoremstyle{definition}
\newtheorem{theorem}{Theorem}[section]
\newtheorem{proposition}[theorem]{Proposition}

\newtheorem{lemma}[theorem]{Lemma}

\theoremstyle{remark}

\newcommand{\jure}[1]{{\color{red}[J: #1]}}
\newcommand{\xhdr}[1]{{\noindent\bfseries #1}.}

\newcommand{\hidecomments}[1]{}
\def\IB{\text{IB}}

\def\GIBCat{\text{GIB-Cat}}
\def\GIBBern{\text{GIB-Bern}}
\def\D{\mathcal{D}}

\title{Graph Information Bottleneck}

\author{%
  Tailin Wu\thanks{Equal contribution},\ \ Hongyu Ren$^*$, Pan Li, Jure Leskovec \\
  Department of Computer Science\\
  Stanford University\\
  \texttt{\{tailin, hyren, panli0, jure\}@cs.stanford.edu} \\
}

\begin{document}

\maketitle

\begin{abstract}
\vspace{-0.1cm}
Representation learning of graph-structured data is challenging because both graph structure and node features carry important information. Graph Neural Networks (GNNs) provide an expressive way to fuse information from network structure and node features. 
However, GNNs are prone to adversarial attacks.
Here we introduce {\em Graph Information Bottleneck (GIB)}, an information-theoretic principle that optimally balances expressiveness and robustness of the learned representation of graph-structured data. Inheriting from the general Information Bottleneck (IB), GIB aims to learn the minimal sufficient representation for a given task by maximizing the mutual information between the representation and the target, and simultaneously constraining the mutual information between the representation and the input data. Different from the general IB, GIB regularizes the \emph{structural} as well as the \emph{feature} information. 
We design two sampling algorithms for structural regularization and instantiate the GIB principle with two new models: GIB-Cat and GIB-Bern, and demonstrate the benefits by evaluating the resilience to adversarial attacks. We show that our proposed models are more robust than state-of-the-art graph defense models. GIB-based models empirically achieve up to 31\% improvement with adversarial perturbation of the graph structure as well as node features.

\end{abstract}

\vspace{-0.1cm}
\section{Introduction}

Representation learning on graphs aims to learn representations of graph-structured data for downstream tasks such as node classification and link prediction~\cite{hamilton2017inductive,kipf2016variational}. Graph representation learning is a challenging task since both node features as well as graph structure carry important information~\cite{kipf2017semi,li2019optimizing}. Graph Neural Networks (GNNs) \cite{kipf2017semi,velickovic2018graph,hamilton2017inductive,chen2018fastgcn,klicpera_predict_2019} have demonstrated impressive performance, by learning to fuse information from both the node features and the graph structure~\cite{xu2018how}. 

Recently, many works have been focusing on developing more powerful GNNs \cite{xu2018how,pmlr-v97-you19b,Pei2020Geom-GCN,maron2019provably,murphy2019relational,chen2019equivalence}, in a sense that they can fit more complex graph-structured data. However, at present GNNs still suffer from a few problems. For example, the features of a neighborhood node can contain non-useful information that may negatively impact the prediction of the current node~\cite{Hou2020Measuring}. Also, GNN's reliance on message passing over the edges of the graph also makes it prone to noise and adversarial attacks that target at the graph structure \cite{zugner2018adversarial,dai2018adversarial}.

\begin{figure}[t]
\begin{minipage}{0.41\textwidth}
\centering
\includegraphics[trim={5.7cm 7.7cm 9.6cm 7.4cm},clip,width=0.95\textwidth]{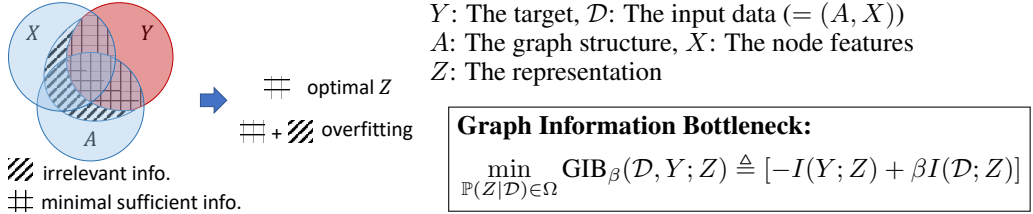}
\end{minipage}
\begin{minipage}{0.59\textwidth}
\flushleft
$Y$: The target, $\mathcal{D}$: The input data ($=(A, X)$) \\
$A$: The graph structure, $X$: The node features \\
$Z$: The representation
\begin{empheq}[box=\fbox]{align*}
    &\textbf{Graph Information Bottleneck: }\\
    &\min_{\mathbb{P}(Z|\mathcal{D})\in \Omega} \text{GIB}_\beta(\mathcal{D},Y;Z) \triangleq {\left[ - I(Y;Z)+\beta I(\mathcal{D};Z) \right]}
\end{empheq}
\end{minipage}
\vspace{-0.0cm}
\caption{\small{Graph Information Bottleneck is to optimize the representation $Z$ to capture the minimal sufficient information within the input data $\mathcal{D}=(A, X)$ to predict the target $Y$. $\mathcal{D}$ includes information from both the graph structure $A$ and node features $X$. When $Z$ contains irrelevant information from either of these two sides, it overfits the data and is prone to adversarial attacks and model hyperparameter change. $\Omega$ defines the search space of the optimal model $\mathbb{P}(Z|\mathcal{D})$. $I(\cdot;\cdot)$ denotes the mutual information~\cite{cover2012elements}}.
}
\label{fig:venn-diag}
\end{figure}

Here we address the above problems and rethink what is a ``good'' representation for graph-structured data. In particular, the Information Bottleneck (IB) \cite{tishby2000information,tishby2015deep} provides a critical principle for representation learning: an optimal representation should contain the \emph{minimal sufficient} information for the downstream prediction task. 
IB encourages the representation to be maximally informative about the target to make the prediction accurate (\emph{sufficient}). On the other hand, IB also discourages the representation from acquiring additional information from the data that is irrelevant for predicting the target (\emph{minimal}). Based on this learning paradigm, the learned model naturally avoids overfitting and becomes more robust to adversarial attacks.

However, extending the IB principle to representation learning on graph-structured data presents two unique challenges. First, previous models that leverage IB assume that the training examples in the dataset are independent and identically distributed (i.i.d.). For graph-structured data, this assumption no longer holds and makes model training in the IB principle  hard. 
Moreover, the structural information is indispensable to represent graph-structured data, but such information is discrete and thus hard to optimize over. How to properly model and extract minimal sufficient information from the graph structure introduces another challenge that has not been yet investigated when designing IB-based models.

We introduce Graph Information Bottleneck (GIB), an information-theoretic principle inherited from IB, adapted for representation learning on graph-structured data. GIB extracts information from both the graph structure and node features and further encourages the information in learned representation to be both minimal and sufficient (Fig.~\ref{fig:venn-diag}). To overcome the challenge induced by non-i.i.d. data, we further leverage local-dependence assumption of graph-structure data to define a more tractable search space $\Omega$ of the optimal $\mathbb{P}(Z|\mathcal{D})$ that follows a Markov chain to hierarchically extract information from both features and structure. To our knowledge, our work provides the first information-theoretic principle 
for supervised representation learning on graph-structured data. 

We also derive variational bounds for GIB, making GIB tractable and amenable for the design and optimization of GNNs. Specifically, we propose a variational upper bound for constraining the information from the node features and graph structure, and a variational lower bound for maximizing the information in the representation to predict the target. 

We demonstrate the GIB principle by applying it to the Graph Attention Networks (GAT) \cite{velickovic2018graph}, where we leverage the attention weights of GAT to sample the graph structure in order to alleviate the difficulty of optimizing and modeling the discrete graph structure. We also design two sampling algorithms based on the categorical distribution and Bernoulli distribution, and propose two models GIB-Cat and GIB-Bern. 
We show that both models consistently improve robustness w.r.t. standard baseline models, and outperform other state-of-the-art defense models.
GIB-Cat and GIB-Bern improve the classification accuracy by up to 31.3\% and 34.0\% under adversarial perturbation, respectively.
\\
Project website and code can be found at \url{http://snap.stanford.edu/gib/}. 
\vspace{-0.1cm}
\section{Preliminaries and Notation}
\vspace{-0.1cm}
\xhdr{Graph Representation Learning}
Consider an undirected attributed graph $G=(V,E,X)$ with $n$ nodes, where $V=[n]=\{1,2,...n\}$ is the node set, $E\subseteq V\times V$ is the edge set and $X\in\mathbb{R}^{n\times f}$ includes the node attributes. Let $A\in \mathbb{R}^{n\times n}$ denote the adjacency matrix of $G$, \textit{i.e.}, $A_{uv} = 1$ if $(u,v)\in E$ or $0$ otherwise. Also, let $d(u,v)$ denote the shortest path distance between two nodes $u,\,v\,(\in V)$ over $A$. Hence our input data can be overall represented as $\mathcal{D} = (A, X)$. 

In this work, we focus on node-level tasks where nodes are associated with some labels $Y\in [K]^n$. Our task is to extract node-level representations $Z_X\in\mathbb{R}^{n\times f'}$ from $\mathcal{D}$ such that $Z_X$ can be further used to predict $Y$. We also use the subscript with a certain node $v\in V$ to denote the affiliation with node $v$. For example, the node representation of $v$ is denoted by $Z_{X,v}$ and its label is denoted by $Y_v$.


\xhdr{Notation}
We do not distinguish the notation of random variables and of their particular realizations if there is no risk of confusion. For any set of random variables $H$, we use $\mathbb{P}(H),\,\mathbb{Q}(H),...$ to denote joint probabilistic distribution functions (PDFs) of the random variables in $H$ under different models. $\mathbb{P}(\cdot)$ corresponds to the induced PDF of the proposed model while $\mathbb{Q}(H)$ and $\mathbb{Q}_i(H)$, $i\in\mathbb{N}$ correspond to some other distributions, typically variational distributions. For discrete random variables, we use generalized PDFs that may contain the Dirac delta functions~\cite{dirac1981principles}. In this work, if not specified, $\mathbb{E}[H]$ means the expectation over all the random variables in $H$ w.r.t. $\mathbb{P}(H)$. Otherwise, we use $\mathbb{E}_{\mathbb{Q}(H)}[H]$ to specify the expectation w.r.t. other distributions denoted by $\mathbb{Q}(H)$. We also use $X_1\perp X_2 | X_3$ to denote that $X_1$ and $X_2$ are conditionally independent given $X_3$. Let $\text{Cat}(\phi)$, $\text{Bernoulli}(\phi)$ denote the categorical distribution and Bernoulli distribution respectively with parameter $\phi\:(\in\mathbb{R}_{\geq 0}^{1\times C})$. For the categorical distribution, $\phi$ corresponds to the probabilities over different categories and thus $\|\phi\|_1 = 1$. For the Bernoulli distribution, we generalize it to high dimensions and assume we have $C$ independent components and each element of $\phi$ is between 0 and 1. Let $\text{Gaussian}(\mu, \sigma^2)$ denote the Gaussian distribution with mean $\mu$ and variance $\sigma^2$. $\mu$ and $\sigma^2$ could be vectors with the same dimension, in which case the Gaussian distribution is with the mean vector $\mu$ and covariance matrix $\Sigma = \text{diag}(\sigma^2)$. Let $\Phi(\cdot: \mu, \sigma^2)$ denote its PDF. We use $[i_1:i_2]$ to slice a tensor w.r.t. indices from $i_1$ to $i_2-1$ of its last dimension.

\vspace{-0.2cm}
\section{Graph Information Bottleneck}
\vspace{-0.1cm}
\subsection{Deriving the Graph Information Bottleneck Principle}

\begin{figure}[t]
\vspace{-0.1cm}
\begin{minipage}{0.745\textwidth}
\centering
\includegraphics[trim={1.5cm 9.4cm 0.8cm 5cm},clip,width=0.99\textwidth]{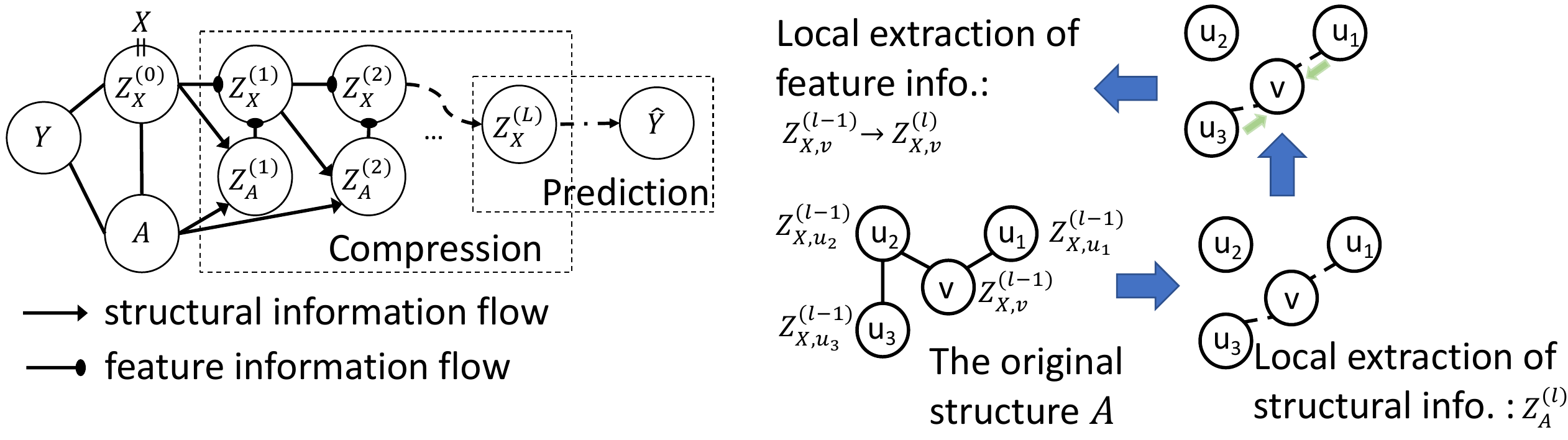}
\hfill\vline\hfill
\end{minipage}
\begin{minipage}{0.245\textwidth}
\centering
\includegraphics[trim={2.3cm 11.7cm 18.7cm 5cm},clip,width=\textwidth]{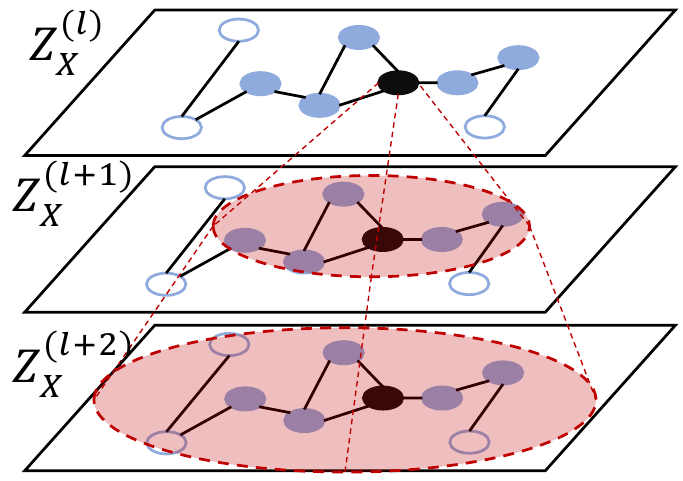}
\end{minipage}
\vspace{-0.1cm}
\caption{\small{Our GIB principle leverages local-dependence assumption. \textbf{(a)} The Markov chain defines the search space $\Omega$ of our GIB principle, of which each step uses a local-dependence assumption to extract information from the structure and node features. The correlation between node representations are established in a hierarchical way: Suppose local dependence appears within 2-hops given the structure $A$. \textbf{(b)} In the graph, given the representations $Z_X^{(l)}$ of the blue nodes and $A$ that conveys the structural information that the blue nodes lie within 2-hops of the black node, the representations $Z_X^{(l+1)}$ are independent between the black node and the white nodes. However, the correlation between them may be established in $Z_X^{(l+2)}$.}}
\label{fig:model-diag}
\vspace{-0.3cm}
\end{figure}

\begin{textblock*}{1cm}(3.9cm,2.4cm) 
   \textbf{(a)}
\end{textblock*}
\begin{textblock*}{1cm}(14.4cm,2.4cm) 
   \textbf{(b)}
\end{textblock*}

In general, the graph information bottleneck (GIB) principle, inheriting from the principle of information bottleneck (IB), requires the node representation $Z_X$ to minimize the information from the graph-structured data $\mathcal{D}$ (compression) and maximize the information to $Y$ (prediction). However, optimization for the most general GIB is challenging because of the correlation between data points. The i.i.d. assumption of data points is typically used to derive variational bounds and make accurate estimation of those bounds to learn IB-based models~\cite{alemi2016deep,poole2019variational}. However, for the graph-structured data $\mathcal{D}$, this is impossible as node features, \textit{i.e.}, different rows of $X$, may be correlated due to the underlying graph structure $A$. To fully capture such correlation, we are not allowed to split the whole graph-structured data $\mathcal{D}$ w.r.t. each node. In practice, we typically have only a large network, which indicates that only one single realization of $\mathbb{P}(\mathcal{D})$ is available. Hence, approximating the optimal $Z_X$ in the general formulation GIB seems impossible without making additional assumptions. 

Here, we rely on a widely accepted \emph{local-dependence} assumption for graph-structured data: Given the data related to the neighbors within a certain number of hops of a node $v$, the data in the rest of the graph will be independent of $v$. We use this assumption to constrain the space $\Omega$ of optimal representations, which leads to a more tractable GIB principle. That is, we assume that the optimal representation follows the Markovian dependence shown in Fig.~\ref{fig:model-diag}. Specifically, $\mathbb{P}(Z_X|\mathcal{D})$ iterates node representations to hierarchically model the correlation. In each iteration $l$, the local-dependence assumption is used: The representation of each node will be refined by incorporating its neighbors w.r.t a graph structure $Z_A^{(l)}$.  Here, $\{Z_A^{(l)}\}_{1\leq l\leq L}$ is obtained by locally adjusting the original graph structure $A$ and essentially controlling the information flow from $A$. 
Finally, we will make predictions based on $Z_X^{(L)}$. Based on this formulation, the objective reduces to the following optimization: 
\begin{align}\label{eq:gib}
\min_{\mathbb{P}(Z_X^{(L)}|\mathcal{D})\in \Omega}\text{GIB}_\beta(\mathcal{D},Y;Z_X^{(L)}) \triangleq {\left[ - I(Y;Z_X^{(L)})+\beta I(\mathcal{D};Z_X^{(L)}) \right]}
\end{align}
where $\Omega$ characterizes the space of the conditional distribution of $Z_X^{(L)}$ given the data $\mathcal{D}$ by following the probabilistic dependence shown in Fig.~\ref{fig:model-diag}. In this formulation, we just need to optimize two series of distributions $\mathbb{P}(Z_X^{(l)}| Z_X^{(l-1)}, Z_A^{(l)})$ and $\mathbb{P}(Z_A^{(l)}| Z_X^{(l-1)}, A)$, $l\in[L]$, which have local dependence between nodes and thus are much easier to be parameterized and optimized.


\textbf{Variational Bounds.} Even using the reduced GIB principle and some proper parameterization of $\mathbb{P}(Z_X^{(l)}| Z_X^{(l-1)}, Z_A^{(l)})$ and $\mathbb{P}(Z_A^{(l)}| Z_X^{(l-1)}, A)$, $l\in[L]$, exact computation of $I(Y;Z_X^{(L)})$ and $I(\mathcal{D};Z_X^{(L)})$ is still intractable. Hence, we need to introduce variational bounds on these two terms, which leads to the final objective to optimize. Note that variational methods are frequently used in model optimization under the traditional IB principle \cite{alemi2016deep}. However, we should be careful to derive these bounds as the data points now are correlated. We introduce a lower bound of $I(Y;Z_X^{(L)})$, which is reproduced from~\cite{poole2019variational,nguyen2010estimating}, and an upper bound of $I(\mathcal{D};Z_X^{(L)})$, as shown in Propositions~\ref{prop:IYZ} and \ref{prop:IXZ}.
\begin{proposition}[The lower bound of $I(Y;Z_X^{(L)})$]\label{prop:IYZ} For any distributions $\mathbb{Q}_1(Y_v|Z_{X,v}^{(L)})$ for $v\in V$ and $\mathbb{Q}_2(Y)$, we have 
\begin{align}\label{eq:iyz}
I(Y;Z_X^{(L)}) \geq 1 +  \mathbb{E}\left[\log\frac{\prod_{v\in V}\mathbb{Q}_1(Y_v|Z_{X,v}^{(L)})}{\mathbb{Q}_2(Y)}\right]  +\mathbb{E}_{\mathbb{P}(Y)\mathbb{P}(Z_{X}^{(L)})}\left[\frac{\prod_{v\in V}\mathbb{Q}_1(Y_v|Z_{X,v}^{(L)})}{\mathbb{Q}_2(Y)}\right] 
\end{align}
\end{proposition}
%
%
\begin{proposition}[The upper bound of $I(\mathcal{D};Z_X^{(L)})$] \label{prop:IXZ}
We choose two groups of indices $S_X, S_A\subset [L]$ such that $\mathcal{D} \perp Z_X^{(L)} | \{Z_X^{(l)}\}_{l\in S_X}\cup \{Z_A^{(l)}\}_{l\in S_A}$ based on the Markovian dependence in Fig.~\ref{fig:model-diag}, and then for any distributions $\mathbb{Q}(Z_X^{(l)})$, $l\in S_X$, and $\mathbb{Q}(Z_A^{(l)})$, $l\in S_A$,
\begin{align}\label{eq:ixz}
&I(\mathcal{D}; Z_X^{(L)}) \leq I(\mathcal{D}; \{Z_X^{(l)}\}_{l\in S_X}\cup \{Z_A^{(l)}\}_{l\in S_A}) \leq \sum_{l\in S_A}\text{AIB}^{(l)} +  \sum_{l\in S_X}\text{XIB}^{(l)}, \text{where}  \\
&\text{AIB}^{(l)} = \mathbb{E}\left[\log \frac{\mathbb{P}(Z_A^{(l)}|A, Z_X^{(l-1)})}{\mathbb{Q}(Z_A^{(l)})} \right], \text{XIB}^{(l)} = \mathbb{E}\left[\log \frac{\mathbb{P}(Z_X^{(l)}|Z_X^{(l-1)}, Z_A^{(l)})}{\mathbb{Q}(Z_X^{(l)})} \right], 
\end{align}
\end{proposition}
The proofs are given in Appendix \ref{app:proof3_1} and \ref{app:proof3_2}. Proposition~\ref{prop:IXZ} indicates that we need to select a group of random variables with index sets $S_X$ and $S_A$ to guarantee the conditional independence between $\mathcal{D}$ and $Z_X^{(L)}$. Note that $S_X$ and $S_A$ that satisfy this condition have the following properties: (1) $S_X\neq \emptyset$,  and (2) suppose the greatest index in $S_X$ is $l$ and then $S_A$ should contain all integers in $[l+1, L]$.

To use GIB, we need to model $\mathbb{P}(Z_A^{(l)}| Z_X^{(l-1)}, A)$ and $\mathbb{P}(Z_X^{(l)}| Z_X^{(l-1)}, Z_A^{(l)})$. Then, we choose some variational distributions $\mathbb{Q}(Z_X^{(l)})$ and $\mathbb{Q}(Z_A^{(l)})$ to estimate the corresponding $\text{AIB}^{(l)}$ and $\text{XIB}^{(l)}$ for regularization, and some  $\mathbb{Q}_1(Y_v|Z_{X,v}^{(L)})$ and $\mathbb{Q}_2(Y)$ to specify the lower bound in Eq. \eqref{eq:iyz}. Then, plugging Eq. \eqref{eq:iyz} and Eq. \eqref{eq:ixz} into the GIB principle (Eq. \eqref{eq:gib}), one obtains an upper bound on the objective to optimize. Note that any model that parameterizes $\mathbb{P}(Z_A^{(l)}| Z_X^{(l-1)}, A)$ and $\mathbb{P}(Z_X^{(l)}| Z_X^{(l-1)}, Z_A^{(l)})$ can use GIB as the objective in training. In the next subsection, we will introduce two instantiations of GIB, which is inspired by GAT~\cite{velickovic2018graph}.   

\subsection{Instantiating the GIB Principle}

The GIB principle can be applied to many GNN models. As an example, we apply it to the Graph Attention Network model~\cite{velickovic2018graph} and present GIB-Cat and GIB-Bern. Algorithm 1 illustrates the base framework of both models with different neighbor sampling methods shown in Algorithm 2 and 3. In each layer, GIB-Cat and GIB-Bern need to first refine the graph structure using the attention weights to obtain $Z_A^{(l)}$ (Step 3) and then refines node representations $Z_X^{(l)}$ by propagating $Z_X^{(l-1)}$ over $Z_A^{(l)}$ (Steps 4-7). 
Concretely, we design two algorithms for neighbor sampling, which respectively use the categorical distribution and the Bernoulli distribution. For the categorical version, we view the attention weights as the parameters of categorical distributions to sample the refined graph structure to extract structural information. We sample $k$ neighbors with replacement from the pool of nodes $V_{vt}$ for each node $v$, where $V_{vt}$ includes the nodes whose shortest-path-distance to $v$ over $A$ is $t$. We use $\mathcal{T}$ as an upper limitation of $t$ to encode the local-dependence assumption of the GIB principle, which also benefits the scalability of the model. For the Bernoulli version, we model each pair of node $v$ and its neighbors independently with a Bernoulli distribution parameterized by the attention weights. Note that here we did not normalize it with the softmax function as in the categorical version, however, we use the sigmoid function to squash it between 0 and 1. Here we do not need to specify the number of neighbors one node sample ($k$ in the categorical version). Step 4 is sum-pooling of the neighbors, and the output will be used to compute the parameters for a Gaussian distribution where the refined node representations will be sampled. Note that we may also use a mechanism similar to multi-head attention~\cite{velickovic2018graph}: We split $\tilde{Z}_X^{(l-1)}$ into different channels w.r.t. its last dimension, perform Steps 2-7 independently for each channel and then concatenate the output of different channels to obtain new $Z_X^{(l)}$. Moreover, when training the model, we adopt reparameterization trick for Steps 3 and 7: Step 3 uses Gumbel-softmax~\cite{jang2016categorical,maddison2016concrete} while Step 7 uses $\hat{Z}_{X,v}^{(l)} = \mu_v^{(l)} + \sigma_v^{(l)}\odot z$ where $z\sim\text{Gaussian}(0, I),\,z\in\mathbb{R}^{1\times f'}$ and $\odot$ is element-wise product.

\begin{wrapfigure}{r}{0.58\textwidth}
\vspace{-0.5cm}
\begin{tabular}{l}
\hline 
\textbf{Algorithm 1: Framework of GIB-Cat and GIB-Bern} \\
\hline
 \textbf{Input:} The dataset $\mathcal{D} = (X, A)$; \\
 $\mathcal{T}$: An integral limitation to impose local dependence; \\
 $k$: The number of neighbors to be sampled. \\
 $\tau$: An element-wise nonlinear rectifier. \\
 \textbf{Initialize:} $Z_X^{(0)} \leftarrow X$; For all  $v\in V,\, t\in [\mathcal{T}]$, \\ 
 construct sets $V_{vt} \leftarrow \{u\in V| d(u,v) = t\}$; \\
 Weights: $a\in \mathbb{R}^{\mathcal{T}\times 4f'}$, $W^{(1)}\in \mathbb{R}^{f\times 2f'}$,\\
 $W^{(l)}\in \mathbb{R}^{f'\times 2f'}$, for $l\in[2,L]$, $W_{\text{out}}\in \mathbb{R}^{f'\times K}$. \\
 \textbf{Output:} $Z_{X}^{(L)}$, $\hat{Y}_v = \text{softmax}(Z_{X,v}^{(L)}W_{\text{out}})$ \\
1. \textbf{For} layers $l = 1, ..., L$ and \textbf{For} $v\in V$, \textbf{do}: \\
2.\quad\vline height 2ex\; $\tilde{Z}_{X,v}^{(l-1)} \leftarrow \tau(Z_{X,v}^{(l-1)})W^{(l)}$ \\
3.\quad\vline height 2.5ex\; $Z_{A,v}^{(l)} \leftarrow \textbf{NeighborSample}(Z_{X}^{l-1}, \mathcal{T}, V_{vt}, a)$\\ 
4.\quad\vline height 2.3ex\; $\bar{Z}_{X,v}^{(l)} \leftarrow \sum_{u\in Z_{A,v}^{(l)}} \tilde{Z}_{X,v}^{(l-1)}$\\
5.\quad\vline height 2.3ex\; $\mu_{v}^{(l)} \leftarrow \bar{Z}_{X,v}^{(l)}[0:f']$ \\
6.\quad\vline height 2.3ex\; $\sigma_{v}^{2(l)} \leftarrow  \text{softplus}(\bar{Z}_{X,v}^{(l)}[f':2f'])$\\
7.\quad\vline height 2.3ex depth 0.5 pt \line(1,0){3.5}\, $Z_{X,v}^{(l)}\sim \text{Gaussian}(\mu_{v}^{(l)}, \sigma_{v}^{2(l)}) $\\ 
\hline 
\end{tabular}
\vspace{-0.5cm}
\end{wrapfigure}

\textbf{Properties.} Different from traditional GNNs, GIB-Cat and GIB-Bern depend loosely on the graph structure since $A$ is only used to decide the potential neighbors for each node, and we perform message passing based on $Z_A$. This property renders our models extremely robust to structural perturbations/attacks where traditional GNNs are sensitive~ \cite{dai2018adversarial,zugner2018adversarial}. Both our models also keep robustness to the feature perturbation that is similar to other IB-based DNN models~\cite{alemi2016deep,fischer2020ceb}. Moreover, the proposed models are invariant to node permutations as we may show that for any permutation matrix $\Pi\in \mathbb{R}^{n\times n}$, with permuting $A\rightarrow A_{\Pi} = \Pi A \Pi^T$, $X\rightarrow X_{\Pi} =\Pi X$, the obtained new node representations $Z_{X, \Pi}^{(L)}$ and $\Pi Z_{X}^{(L)}$ share the same distribution (proof in Appendix~\ref{app:perm-inv}). Permutation invariance is known to be important for structural representation learning~\cite{chen2019equivalence}.

\resizebox{0.5\columnwidth}{!}{%
\begin{tabular}{l}
\hline 
\textbf{Algorithm 2: NeighborSample (categorical)} \\
\hline
 \textbf{Input:} $Z_{X}^{l}, \mathcal{T}, V_{vt}, a$, as defined in Alg. 1; \\
 \textbf{Output:} $Z_{A,v}^{(l+1)}$\\
1.\textbf{For} $t\in  [\mathcal{T}]$, \textbf{do}: \\ 
2.\quad\vline height 2.3ex depth 0.5 pt \line(1,0){3.5}\, $\phi_{vt}^{(l)} \leftarrow \text{softmax}(\{ (\tilde{Z}_{X,v}^{(l-1)}\oplus\tilde{Z}_{X,u}^{(l-1)})a^T\}_{u\in V_{vt}})$\\
3. $Z_{A,v}^{(l+1)} \leftarrow \cup_{t=1}^{\mathcal{T}}\{u \in V_{vt}| u\stackrel{\text{iid}}{\sim} \text{Cat}(\phi_{vt}^{(l)}),\,\text{$k$ times}\}$\\
\hline 
\end{tabular}
}
\resizebox{0.5\columnwidth}{!}{%
\begin{tabular}{l}
\hline 
\textbf{Algorithm 3: NeighborSample (Bernoulli)} \\
\hline
 \textbf{Input:} $Z_{X}^{l}, \mathcal{T}, V_{vt}, a$, as defined in Alg. 1; \\
 \textbf{Output:} $Z_{A,v}^{(l+1)}$\\
1.\textbf{For} $t\in  [\mathcal{T}]$, \textbf{do}: \\ 
2.\quad\vline height 2.3ex depth 0.5 pt \line(1,0){3.5}\, $\phi_{vt}^{(l)} \leftarrow \text{sigmoid}(\{ (\tilde{Z}_{X,v}^{(l-1)}\oplus\tilde{Z}_{X,u}^{(l-1)})a^T\}_{u\in V_{vt}})$\\
3. $Z_{A,v}^{(l+1)} \leftarrow \cup_{t=1}^{\mathcal{T}}\{u \in V_{vt}| u\stackrel{\text{iid}}{\sim} \text{Bernoulli}(\phi_{vt}^{(l)})\}$\\
\hline 
\end{tabular}
}

\textbf{Objective for training.} To optimize the parameters of the model, we need to specify the bounds for $I(\mathcal{D}; Z_X^{(L)})$ as in Eq.~\eqref{eq:ixz} and $I(Y; Z_X^{(L)})$ as in Eq.~\eqref{eq:iyz}, and further compute the bound of the GIB objective in Eq.~\eqref{eq:gib}. To characterize $\text{AIB}^{(l)}$ in Eq.~\eqref{eq:ixz}, we assume $\mathbb{Q}(Z_A^{(l)})$ is a non-informative distribution \cite{jang2016categorical,maddison2016concrete}. Specifically, we use the uniform distribution for the categorical version: $Z_A \sim \mathbb{Q}(Z_A)$, $Z_{A,v} =\cup_{t=1}^\mathcal{T}\{u \in V_{vt}| u\stackrel{\text{iid}}{\sim} \text{Cat}(\frac{\mathbf{1}}{|V_{vt}|})\}$ and $Z_{A,v}\perp Z_{A,u}$ if $v\neq u$; and we also adopt a non-informative prior for the Bernoulli version: $Z_{A,v} =\cup_{t=1}^\mathcal{T}\{u \in V_{vt}| u\stackrel{\text{iid}}{\sim} \text{Bernoulli}(\mathbf{\alpha})\}$, where $\alpha\in(0,1)$ is a hyperparameter. The difference is that, unlike the categorical distribution, we have an additional degree of freedom provided by $\alpha$. After the model computes $\phi_{vt}^{(l)}$ according to Step 4, we get an empirical estimation of $\text{AIB}^{(l)}$:
\begin{align*}
    \widehat{\text{AIB}}^{(l)} = \mathbb{E}_{\mathbb{P}(Z_A^{(l)}|A, Z_X^{(l-1)})}\left[\log \frac{\mathbb{P}(Z_A^{(l)}|A, Z_X^{(l-1)})}{\mathbb{Q}(Z_A^{(l)})}\right],
\end{align*}

which is instantiated as follows for the two versions,
\begin{align*}
    \widehat{\text{AIB}_\text{C}}^{(l)} &=  \sum_{v\in V,t\in [\mathcal{T}]}\text{KL}(\text{Cat}(\phi_{vt}^{(l)})||\text{Cat}(\frac{\mathbf{1}}{|V_{vt}|})) \\
    \widehat{\text{AIB}_\text{B}}^{(l)} &=\sum_{v\in V,t\in [\mathcal{T}]}\text{KL}(\text{Bernoulli}(\phi_{vt}^{(l)})||\text{Bernoulli}(\alpha))
\end{align*}



To estimate $\text{XIB}^{(l)}$,  we set $\mathbb{Q}(Z_X^{(l)})$ as a mixture of Gaussians with learnable parameters \cite{dilokthanakul2016deep}. Specifically, for any node $v$, $Z_X\sim \mathbb{Q}(Z_X)$, we set $Z_{X,v} \sim \sum_{i=1}^{m} w_i\text{Gaussian}(\mu_{0,i}, \sigma_{0,i}^2)$ where $w_i, \mu_{0,i}, \sigma_{0,i}$ are learnable parameters shared by all nodes and $Z_{X,v}\perp Z_{X,u}$ if $v\neq u$. We estimate $\text{XIB}^{(l)}$ by using the sampled $Z_X^{(l)}$:
\begin{align*}
    \widehat{\text{XIB}}^{(l)} = \log \frac{\mathbb{P}(Z_X^{(l)}|Z_X^{(l-1)}, Z_A^{(l)})}{\mathbb{Q}(Z_X^{(l)})} = \sum_{v\in V} \left[\log \Phi(Z_{X,v}^{(l)}; \mu_v, \sigma^2_v) - \log (\sum_{i=1}^m w_i\Phi(Z_{X,v}^{(l)}; \mu_{0,i}, \sigma^2_{0,i})) \right].
\end{align*}

Therefore, in practice, we may select proper sets of indices $S_X$, $S_A$ that satisfy the condition in Proposition~\ref{prop:IXZ} and use substitution \vspace{-0.1cm}
\begin{align}\label{eq:prac-ixz}
    I(\mathcal{D}; Z_X^{(L)}) \rightarrow \sum_{l\in S_A}\widehat{\text{AIB}}^{(l)}+ \sum_{l\in S_X}\widehat{\text{XIB}}^{(l)}
\end{align}
To characterize Eq.~\eqref{eq:iyz}, we may simply set $\mathbb{Q}_2(Y) = \mathbb{P}(Y)$ and $\mathbb{Q}_1(Y_v|Z_{X,v}^{(L)}) = \text{Cat}(Z_{X,v}^{(L)}W_{\text{out}})$. Then, the RHS of Eq.~\eqref{eq:iyz} reduces to the cross-entropy loss by ignoring constants, \textit{i.e.},  
\begin{align}\label{eq:prac-iyz}
    I(Y;Z_X^{(L)}) \rightarrow -\sum_{v\in V} \text{Cross-Entropy}(Z_{X,v}^{(L)}W_{\text{out}}; Y_v)
\end{align}

Other choices of $\mathbb{Q}_2(Y)$ may also be adopted and yield the contrastive loss \cite{oord2018representation,poole2019variational} (Appendix~\ref{app:cons-loss}). However, in our case, we use the simplest setting to illustrate the benefit of the GIB principle. Plugging Eq.~\eqref{eq:prac-ixz} and Eq.~\eqref{eq:prac-iyz} into Eq.~\eqref{eq:gib}, we obtain the objective to train our models.

\textbf{Other Formalizations of the GIB Principle.} There are also other alternative formalizations of the GIB principle, especially when modeling $\mathbb{P}(Z_A^{(l)}| Z_X^{(l-1)}, A)$. 
Generally speaking, any node-pair representations, such as messages over edges in MPNN~\cite{gilmer2017neural}, can be leveraged to sample structures. Applying the GIB principle to other architectures is a promising direction for future investigation.

\vspace{-0.1cm}
\section{Related Work}
\vspace{-0.05cm}

GNNs learn node-level representations through message passing and aggregation from neighbors \cite{gilmer2017neural,kipf2017semi,li2018adaptive,xu2018representation,hamilton2017inductive}. Several previous works further incorporate the attention mechanism to adaptively learn the correlation between a node and its neighbor \cite{velickovic2018graph,zhang2018gaan}. \hidecomments{or leverage random walks and pagerank algorithms \cite{klicpera_predict_2019}. To deal with large graphs, several methods propose to perform layer-wise \cite{hamilton2017inductive,chen2018fastgcn,huang2018adaptive} or graph-wise sampling \cite{chiang2019cluster,zeng2019graphsaint} and also increase the number of GCN layers for more expressiveness \cite{Rong2020DropEdge,li2019deepgcns}. A more extensive literature on graph neural networks can be found at \cite{chami2020machine}.}
Recent literature shows that representations learned by GNNs are far from robust and can be easily attacked by malicious manipulation on either features or structure \cite{dai2018adversarial,zugner2018adversarial}. Accordingly, several defense models are proposed to increase the robustness by injecting random noise in the representations \cite{zhu2019robust}, removing suspicious and uninformative edges \cite{wu2019adversarial}, low-rank approximation of the adjacency matrix \cite{entezari2020all}, additional hinge loss for certified robustness \cite{zugner2019certifiable}. In contrast, even though not specifically designed against adversarial attacks, our model learns robust representations via the GIB principle that naturally defend against attacks. Moreover, none of those defense models has theoretical foundations except \cite{zugner2019certifiable} that uses tools of robust optimization instead of information theory.  


Recently several works have applied contrastive loss \cite{oord2018representation} as a regularizer for GNNs. The idea is to increase the score for positive samples while decrease the score for negative samples. This can be further formulated as a mutual information maximization term that aims to maximize the mutual information between representations of nodes and their neighbor patches \cite{dgi}, between representations of sub-structures and the hidden feature vectors \cite{gmi}, between representations of graphs and their sub-structures \cite{sun2019infograph}. In contrast, our model focuses on the compression of node features and graph structure while at the same time improves prediction, which is orthogonal to these previous works on unsupervised representation learning with information maximization.

\hidecomments{
Another line of related works is deep representation learning with the IB principle. DVIB \cite{alemi2016deep} first applies IB \cite{tishby2000information} to deep neural networks with efficient optimization on the variational bound using the reparameterization trick \cite{kingma2013auto}, and shows increased robustness of learned representations. Other works apply IB to various domains, including stabilization of generative adversarial nets \cite{vdb} and learning disentangled representations \cite{betavae}. We refer the reader to \cite{goldfeld2020information} for a more extensive survey of IB. The difference between previous works on IB and our work here is that we develop information-theoretic modeling of both feature and structure and their fusion on graph-structured data.
}

Another line of related work is representation learning with the IB principle. DVIB \cite{alemi2016deep} first applies IB \cite{tishby2000information} to deep neural networks, and shows increased robustness of learned representations. Other methods apply IB to various domains \cite{vdb,betavae}. 
The difference is that we develop information-theoretic modeling of feature, structure and their fusion on graph-structured data. 
Furthermore, several works on GNNs \cite{dgi,gmi,sun2019infograph} leverage information maximization \cite{hjelm2018learning} for unsupervised learning. However, we focus on learning robust representations by \emph{controlling} the information in supervised learning setting.

\vspace{-0.1cm}
\section{Experiments}
\vspace{-0.1cm}
The goal of our experiments is to test whether GNNs trained with the GIB objective are more robust and reliable. Specifically, we consider the following two questions: (1) Boosted by GIB, does \GIBCat{} and \GIBBern{} learn more robust representations than GAT to defend against attacks? (2) How does each component of GIB contribute to such robustness, especially, to controlling the information from one of the two sides --- the structure and node features?  

We compare \GIBCat{} and \GIBBern{} with baselines including GCN \cite{kipf2017semi} and GAT \cite{velickovic2018graph}, the most relevant baseline as \GIBCat{} and \GIBBern{} are to impose the GIB principle over GAT. 
In addition, we consider two state-of-the-art graph defense models specifically designed against adversarial attacks: GCNJaccard \cite{wu2019adversarial} that pre-processes the graph by deleting the edges between nodes with low feature similarity, and Robust GCN (RGCN) \cite{zhu2019robust} that uses Gaussian reparameterization for node features and variance-based attention. Note that RGCN essentially includes the term XIB (Eq.~\eqref{eq:ixz}) to control the information of node features while it does not have the term AIB (Eq.~\eqref{eq:ixz}) 
to control the structural information. For GCNJaccard and RGCN, we perform extensive hyperparameter search as detailed in Appendix \ref{app:implementation_defense_baseline}. For \GIBCat{} and \GIBBern{}, we keep the same architectural component as GAT, and for the additional hyperparameters $k$ and $\mathcal{T}$ (Algorithm 1, 2 and 3), 
we search $k\in\{2,3\}$ and $\mathcal{T}\in\{1,2\}$ for each experimental setting and report the better performance. Please see Appendix \ref{app:implementation_models} for more details.

We use three citation benchmark datasets: Cora, Pubmed and Citeseer \cite{sen2008collective}, in our evaluation. In all experiments, we follow the standard \emph{transductive} node classification setting and standard train-validation-test split as GAT~\cite{velickovic2018graph}. The summary statistics of the datasets and their splitting are shown in Table \ref{table:datasets} in Appendix \ref{app:datasets}.
For all experiments, we perform the experiments over 5 random initializations 
and report average performance. We always use F1-micro as the validating metric to train our model.

\begin{table}[t]
\caption{
Average classification accuracy (\%) for the targeted nodes under direct attack. Each number is the average accuracy for the 40 targeted nodes for 5 random initialization of the experiments. Bold font denotes top two models.
}
\label{table:defense_adverserial}
\resizebox{\textwidth}{!}{
\begin{tabular}{|l|l|c|cccc|cccc|}
\hline
  & \multicolumn{1}{c|}{\multirow{2}{*}{\textbf{Model}}} & \textbf{Clean (\%)}     & \multicolumn{4}{c|}{\textbf{Evasive (\%)}}                                                        & \multicolumn{4}{c|}{\textbf{Poisoning (\%)}}                                                      \\ \cline{3-11} 
                                   & \multicolumn{1}{c|}{}                                & \multicolumn{1}{l|}{}   & \textbf{1}             & \textbf{2}             & \textbf{3}             & \textbf{4}             & \textbf{1}             & \textbf{2}             & \textbf{3}             & \textbf{4}             \\ \hline
\parbox[t]{2mm}{\multirow{5}{*}{\rotatebox[origin=c]{90}{\textbf{Cora}}}}     & GCN                                                  & \textbf{80.0}\scriptsize{$\pm$7.87}  & 51.5\scriptsize{$\pm$4.87}          & 38.0\scriptsize{$\pm$6.22}          & 31.0\scriptsize{$\pm$2.24}          & 26.0\scriptsize{$\pm$3.79}          & 47.5\scriptsize{$\pm$7.07}          & 39.5\scriptsize{$\pm$2.74}          & 30.0\scriptsize{$\pm$5.00}          & 26.5\scriptsize{$\pm$3.79}          \\ 
                                   & GCNJaccard                                           & 75.0\scriptsize{$\pm$5.00}           & 48.5\scriptsize{$\pm$6.75}          & 36.0\scriptsize{$\pm$6.51}          & 32.0\scriptsize{$\pm$3.25}          & 30.0\scriptsize{$\pm$3.95}          & 47.0\scriptsize{$\pm$7.37}          & 38.0\scriptsize{$\pm$6.22}          & 33.5\scriptsize{$\pm$3.79}          & 28.5\scriptsize{$\pm$3.79}          \\ 
                                   & RGCN                                                 & \textbf{80.0}\scriptsize{$\pm$4.67}  & 49.5\scriptsize{$\pm$6.47}          & 36.0\scriptsize{$\pm$5.18}          & 30.5\scriptsize{$\pm$3.25}          & 25.5\scriptsize{$\pm$2.09}          & 46.5\scriptsize{$\pm$5.75}          & 35.5\scriptsize{$\pm$3.70}           & 29.0\scriptsize{$\pm$3.79}          & 25.5\scriptsize{$\pm$2.73}          \\ 
                                   & GAT                                                  & 77.8\scriptsize{$\pm$3.97}           & 48.0\scriptsize{$\pm$8.73}          & 39.5\scriptsize{$\pm$5.70}          & 36.5\scriptsize{$\pm$5.48}          & 32.5\scriptsize{$\pm$5.30}           & 50.5\scriptsize{$\pm$5.70}          & 38.0\scriptsize{$\pm$5.97}          & 33.5\scriptsize{$\pm$2.85}          & 26.0\scriptsize{$\pm$3.79}          \\ \cline{2-11} 
                                   & \textbf{GIB-Cat}                                     & 77.6\scriptsize{$\pm$2.84} & \textbf{63.0}\scriptsize{$\pm$4.81} & \textbf{52.5}\scriptsize{$\pm$3.54} & \textbf{44.5}\scriptsize{$\pm$5.70} & \textbf{36.5}\scriptsize{$\pm$6.75} & \textbf{60.0}\scriptsize{$\pm$6.37} & \textbf{50.0}\scriptsize{$\pm$2.50}  & \textbf{39.5}\scriptsize{$\pm$5.42} & \textbf{30.0}\scriptsize{$\pm$3.95} \\
                       
                        & \textbf{GIB-Bern}
                        &
                        78.4\scriptsize{$\pm$4.07} & \textbf{64.0}\scriptsize{$\pm$5.18} & \textbf{51.5}\scriptsize{$\pm$4.54} & \textbf{43.0}\scriptsize{$\pm$3.26} & \textbf{37.5}\scriptsize{$\pm$3.95} & \textbf{61.5}\scriptsize{$\pm$4.18} & \textbf{46.0}\scriptsize{$\pm$4.18}  & \textbf{36.5}\scriptsize{$\pm$4.18} & \textbf{31.5}\scriptsize{$\pm$2.85} \\
                                   \hline\hline
\parbox[t]{1mm}{\multirow{5}{*}{\rotatebox[origin=c]{90}{\textbf{Pubmed}}}}   & GCN                                                  & 82.6\scriptsize{$\pm$6.98}           & 39.5\scriptsize{$\pm$4.81}          & 32.0\scriptsize{$\pm$4.81}          & 31.0\scriptsize{$\pm$5.76}          & 31.0\scriptsize{$\pm$5.76} & 36.0\scriptsize{$\pm$4.18}          & 32.5\scriptsize{$\pm$6.37}          & 31.0\scriptsize{$\pm$5.76}          & \textbf{28.5}\scriptsize{$\pm$5.18} \\ 
                                   & GCNJaccard                                           & 82.0\scriptsize{$\pm$7.15}           & 37.5\scriptsize{$\pm$5.30}          & 31.5\scriptsize{$\pm$5.18}          & 30.0\scriptsize{$\pm$3.95}          & 30.0\scriptsize{$\pm$3.95}          & 36.0\scriptsize{$\pm$3.79}          & 32.5\scriptsize{$\pm$4.67}          & 31.0\scriptsize{$\pm$4.87}          & \textbf{28.5}\scriptsize{$\pm$4.18} \\ 
                                   & RGCN                                                 & 79.0\scriptsize{$\pm$5.18}           & 39.5\scriptsize{$\pm$5.70}          & 33.0\scriptsize{$\pm$4.80}          & 31.5\scriptsize{$\pm$4.18}         & 30.0\scriptsize{$\pm$5.00}          & 38.5\scriptsize{$\pm$4.18}          & 31.5\scriptsize{$\pm$2.85}          & 29.5\scriptsize{$\pm$3.70}          & 27.0\scriptsize{$\pm$3.70}          \\ 
                                   & GAT                                                  & 78.6\scriptsize{$\pm$6.70}           & 41.0\scriptsize{$\pm$8.40}          & 33.5\scriptsize{$\pm$4.18}          & 30.5\scriptsize{$\pm$4.47}          & 31.0\scriptsize{$\pm$4.18} & 39.5\scriptsize{$\pm$3.26}          & 31.0\scriptsize{$\pm$4.18}          & 30.0\scriptsize{$\pm$3.06}          & 25.5\scriptsize{$\pm$5.97}          \\ \cline{2-11} 
                                   & \textbf{GIB-Cat}                                     & \textbf{85.1}\scriptsize{$\pm$6.90}  & \textbf{72.0}\scriptsize{$\pm$3.26} & \textbf{51.0}\scriptsize{$\pm$5.18} & \textbf{37.5}\scriptsize{$\pm$5.30} & \textbf{31.5}\scriptsize{$\pm$4.18} & \textbf{71.0}\scriptsize{$\pm$4.87} & \textbf{48.0}\scriptsize{$\pm$3.26} & \textbf{37.5}\scriptsize{$\pm$1.77} & \textbf{28.5}\scriptsize{$\pm$2.24} \\
                            & \textbf{GIB-Bern}                                     & \textbf{86.2}\scriptsize{$\pm$6.54}  & \textbf{76.0}\scriptsize{$\pm$3.79} & \textbf{50.5}\scriptsize{$\pm$4.11} & \textbf{37.5}\scriptsize{$\pm$3.06} & \textbf{31.5}\scriptsize{$\pm$1.37} & \textbf{72.5}\scriptsize{$\pm$4.68} & \textbf{48.0}\scriptsize{$\pm$2.74} & \textbf{36.0}\scriptsize{$\pm$2.85} & 26.5\scriptsize{$\pm$2.85} \\        
                                   
                                   \hline\hline
\parbox[t]{1mm}{\multirow{5}{*}{\rotatebox[origin=c]{90}{\textbf{Citeseer}}}} & GCN                                                  & 71.8\scriptsize{$\pm$6.94}           & 42.5\scriptsize{$\pm$7.07}          & 27.5\scriptsize{$\pm$6.37}          & 18.0\scriptsize{$\pm$3.26}          & 15.0\scriptsize{$\pm$2.50}          & 29.0\scriptsize{$\pm$7.20}          & 20.5\scriptsize{$\pm$1.12}          & \textbf{17.5}\scriptsize{$\pm$1.77} & \textbf{13.0}\scriptsize{$\pm$2.09}          \\ 
                                   & GCNJaccard                                           & \textbf{72.5}\scriptsize{$\pm$9.35}           & 41.0\scriptsize{$\pm$6.75}          & 32.5\scriptsize{$\pm$3.95}          & 20.5\scriptsize{$\pm$3.70}          & 13.0\scriptsize{$\pm$1.11}          & \textbf{42.5}\scriptsize{$\pm$5.86} & \textbf{30.5}\scriptsize{$\pm$5.12} & \textbf{17.5}\scriptsize{$\pm$1.76} & \textbf{14.0}\scriptsize{$\pm$1.36} \\ 
                                   & RGCN                                                 & \textbf{73.5}\scriptsize{$\pm$8.40} & 41.5\scriptsize{$\pm$7.42}         & 24.5\scriptsize{$\pm$6.47}          & 18.5\scriptsize{$\pm$6.52}          & 13.0\scriptsize{$\pm$1.11}          & 31.0\scriptsize{$\pm$5.48}          & 19.5\scriptsize{$\pm$2.09}          & 13.5\scriptsize{$\pm$2.85}          & 5.00\scriptsize{$\pm$1.77}          \\ 
                                   & GAT                                                  & 72.3\scriptsize{$\pm$8.38}           & \textbf{49.0}\scriptsize{$\pm$9.12} & 33.0\scriptsize{$\pm$5.97} & 22.0\scriptsize{$\pm$4.81}          & 18.0\scriptsize{$\pm$3.26}          & \textbf{38.0}\scriptsize{$\pm$5.12}          & \textbf{23.5}\scriptsize{$\pm$4.87}          & 16.5\scriptsize{$\pm$4.54}          & 12.0\scriptsize{$\pm$2.09}          \\ \cline{2-11} 
                                   & \textbf{GIB-Cat}                                     & 68.6\scriptsize{$\pm$4.90}           & \textbf{51.0}\scriptsize{$\pm$4.54}          & \textbf{39.0}\scriptsize{$\pm$4.18}          & \textbf{32.0}\scriptsize{$\pm$4.81} & \textbf{26.5}\scriptsize{$\pm$4.54} & 30.0\scriptsize{$\pm$9.19}          & 14.0\scriptsize{$\pm$5.76}          & 9.50\scriptsize{$\pm$3.26}          & 6.50\scriptsize{$\pm$2.24}          \\
                            & \textbf{GIB-Bern}                                     & 71.8\scriptsize{$\pm$5.03}           & \textbf{49.0}\scriptsize{$\pm$7.42}          & \textbf{37.5}\scriptsize{$\pm$7.71}          & \textbf{32.5}\scriptsize{$\pm$4.68} & \textbf{23.5}\scriptsize{$\pm$7.42} & 35.0\scriptsize{$\pm$6.37}          & 19.5\scriptsize{$\pm$4.81}          & 11.5\scriptsize{$\pm$3.79}          & 6.00\scriptsize{$\pm$2.85}          \\         
                                   \hline
\end{tabular}
}
\vspace{-0.1cm}
\end{table}

\begin{table}[t]
\vskip -0.1in
\caption{
Average classification accuracy (\%) for the ablations of GIB-Cat and GIB-Bern on Cora dataset.}
\label{table:ablation_cora}
\resizebox{\textwidth}{!}{
\begin{tabular}{|l|c|cccc|cccc|}
\hline
\multirow{2}{*}{\textbf{Model}} & \textbf{Clean (\%)}                              & \multicolumn{4}{c|}{\textbf{Evasive (\%)}}                                                                                                                                                 & \multicolumn{4}{c|}{\textbf{Poisoning (\%)}}                                                                                                                                               \\ \cline{2-10} 
                                 &                                             & \textbf{1}                                  & \textbf{2}                                  & \textbf{3}                                  & \textbf{4}                                  & \textbf{1}                                  & \textbf{2}                                  & \textbf{3}                                  & \textbf{4}                                  \\ \hline

\textbf{XIB}   & 76.3\scriptsize{$\pm$2.90} & 57.0\scriptsize{$\pm$5.42} & 47.5\scriptsize{$\pm$7.50} & 39.5\scriptsize{$\pm$6.94} & 33.0\scriptsize{$\pm$3.71} & 54.5\scriptsize{$\pm$2.09} & 41.0\scriptsize{$\pm$3.79}  & 36.0\scriptsize{$\pm$5.18} & 31.0\scriptsize{$\pm$4.54} \\ 
\textbf{AIB-Cat}   & 78.7\scriptsize{$\pm$4.95} & 62.5\scriptsize{$\pm$5.86} & 51.5\scriptsize{$\pm$5.18} & 43.0\scriptsize{$\pm$3.26} & 36.0\scriptsize{$\pm$3.35} & 60.5\scriptsize{$\pm$3.26} & 47.5\scriptsize{$\pm$5.00} & 36.0\scriptsize{$\pm$3.35} & 31.5\scriptsize{$\pm$6.27} \\
\textbf{AIB-Bern}   & 79.9\scriptsize{$\pm$3.78} & 64.0\scriptsize{$\pm$4.50} & 51.5\scriptsize{$\pm$6.50} & 42.0\scriptsize{$\pm$5.40} & 37.0\scriptsize{$\pm$5.70} & 58.5\scriptsize{$\pm$3.80} & 46.0\scriptsize{$\pm$4.50} & 39.0\scriptsize{$\pm$4.20} & 30.0\scriptsize{$\pm$3.10} \\
\hline
\textbf{GIB-Cat}                                     & 77.6\scriptsize{$\pm$2.84} & 63.0\scriptsize{$\pm$4.81} & 52.5\scriptsize{$\pm$3.54} & 44.5\scriptsize{$\pm$5.70} & 36.5\scriptsize{$\pm$6.75} & 60.0\scriptsize{$\pm$6.37} & 50.0\scriptsize{$\pm$2.50}  & 39.5\scriptsize{$\pm$5.42} & 30.0\scriptsize{$\pm$3.95} \\ 
\textbf{GIB-Bern}
    &
    78.4\scriptsize{$\pm$4.07} & 64.0\scriptsize{$\pm$5.18} & 51.5\scriptsize{$\pm$4.54} & 43.0\scriptsize{$\pm$3.26} & 37.5\scriptsize{$\pm$3.95} & 61.5\scriptsize{$\pm$4.18} & 46.0\scriptsize{$\pm$4.18}  & 36.5\scriptsize{$\pm$4.18} & 31.5\scriptsize{$\pm$2.85} \\
\hline
\end{tabular}
}
\vskip -0.1in
\end{table}
\subsection{Robustness Against Adversarial Attacks}
\label{sec:robustness_adversary}
In this experiment, we compare the robustness of different models against adversarial attacks. We use Nettack \cite{zugner2018adversarial}, a strong targeted attack technique on graphs that attacks a target node by flipping the edge or node features. We evaluate the models on both evasive and poisoning settings, \textit{i.e.} the attack happens after or before the model is trained, respectively.  We follow the setting of Nettack \cite{zugner2018adversarial}:
for each dataset, select (i) 10 nodes with highest margin of classification, \textit{i.e.} they are clearly correctly classified, (ii) 10 nodes with lowest margin but still correctly classified and (iii) 20 more nodes randomly, 
where for each target node, we train a different model for evaluation. We report the classification accuracy of these 40 targeted nodes.
We enumerate the number of perturbations from 1 to 4, where each perturbation denotes a flipping of a node feature or an addition or deletion of an edge. Since Nettack can only operate on Boolean features, we binarize the node features before training. 

Table \ref{table:defense_adverserial} shows the results. We see that compared with GAT, \GIBCat{} improves the classification accuracy by an average of 8.9\% and 14.4\% in Cora and Pubmed, respectively, and \GIBBern{} improves the classification accuracy by an average of 8.4\% and 14.6\% in Cora and Pubmed, respectively, which demonstrates the effectiveness of the GIB principle to improve the robustness of GNNs. Remarkably, when the number of perturbation is 1, \GIBCat{} and \GIBBern{} boost accuracy over GAT (as well as other models) by 31.3\% and 34.0\% in Pubmed, respectively. \GIBCat{} also outperforms GCNJaccard and RGCN by an average of 10.3\% and 12.3\% on Cora (For \GIBBern{}, it is 9.8\% and 11.7\%), and by an average of 15.0\% and 14.6\% on Pubmed (For \GIBBern{}, it is 15.2\% and 14.8\%), although \GIBCat{} and \GIBBern{} are not intentionally designed to defend attacks. For Citeseer, \GIBCat{} and \GIBBern{}'s performance are worse than GCNJaccard in the poisoning setting. This is because Citeseer has much more nodes with very few degrees, even fewer than the number of specified perturbations, as shown in Table \ref{tab:analysis} in Appendix \ref{app:citeseer_analysis}. In this case, the most effective attack is to connect the target node to a node from a different class with very different features, which exactly matches the assumption used by GCNJaccard \cite{wu2019adversarial}. GCNJaccard proceeds to delete edges with dissimilar node features, resulting in the best performance in Citeseer. However, GIB does not depend on such a restrictive assumption. More detailed analysis is at Appendix~\ref{app:citeseer_analysis}.

\textbf{Ablation study.} To see how different components of GIB contribute to the performance, we perform ablation study on Cora, as shown in Table \ref{table:ablation_cora}. Here, we use AIB-Cat and AIB-Bern to denote the models that only sample structures with $\widehat{\text{AIB}}$ (Eq. \eqref{eq:prac-ixz}) in the objective (whose NeighborSample() function is identical to that of GIB-Cat and GIB-Bern, respectively), and use XIB to denote the model that only samples node representations with $\widehat{\text{XIB}}$ (Eq. \eqref{eq:prac-ixz}) in the objective. We see that the AIB (structure) contributes significantly to the improvement of GIB-Cat and GIB-Bern, and on average, AIB-Cat (AIB-Bern) only underperforms GIB-Cat (GIB-Bern) by 0.9\% (0.4\%). 
The performance gain is due to the attacking style of Nettack, as the most effective attack is typically via structural perturbation \cite{zugner2018adversarial}, as is also confirmed in Appendix \ref{app:citeseer_analysis}. Therefore, next we further investigate the case that only perturbation on node features is available.


\begin{table}[t]
\caption{
Classification F1-micro (\%) for the trained models with increasing additive feature noise. Bold font denotes top 2 models.}
\label{table:feature_noise}
\begin{center}
\scalebox{0.8}{
\begin{tabular}{|c|c|c|c|c|}
\hline

\multirow{2}{*}{\textbf{Dataset}}           & \multirow{2}{*}{\textbf{Model}} & \multicolumn{3}{c|}{\textbf{Feature noise ratio ($\lambda$)}}                                                                                                                                                                                                                                 \\ \cline{3-5} 
                                   &                                                                                                   & \textbf{0.5}                                          & \textbf{1}                                        & \textbf{1.5}                                           \\ \hline

\multirow{6}{*}{\textbf{Cora}}     & GCN                       & 64.0\scriptsize{$\pm$2.05}                 & 41.3\scriptsize{$\pm$2.05}   & 31.4\scriptsize{$\pm$2.81}        \\ 
                                  & GCNJaccard                 & 61.1\scriptsize{$\pm$2.18}                  & 41.2\scriptsize{$\pm$2.28}      &  31.8\scriptsize{$\pm$2.63}  \\ 
                                  & RGCN                       & 57.7\scriptsize{$\pm$2.27}                 & 39.1\scriptsize{$\pm$1.58}  &  29.6\scriptsize{$\pm$2.47}       \\ 
                                   & GAT                 & 62.5\scriptsize{$\pm$1.97}           & 41.7\scriptsize{$\pm$2.32}          & 29.8\scriptsize{$\pm$2.98}          \\  \cline{2-5}  
                                  & AIB-Cat & 67.9\scriptsize{$\pm$2.65}           & \textbf{49.6}\scriptsize{$\pm$5.35}          & \textbf{38.4}\scriptsize{$\pm$5.06}          \\
                                  
                                   & AIB-Bern & \textbf{68.8}\scriptsize{$\pm$1.85}           & 49.0\scriptsize{$\pm$2.87}          & 37.1\scriptsize{$\pm$4.47}          \\
                           
                                   & \textbf{GIB-Cat}           & 67.1\scriptsize{$\pm$2.21}  & 49.1\scriptsize{$\pm$3.67}  & 37.5\scriptsize{$\pm$4.76} \\ 
                                     & \textbf{GIB-Bern}           & \textbf{69.0}\scriptsize{$\pm$1.91}  & \textbf{51.3}\scriptsize{$\pm$2.62}  & \textbf{38.9}\scriptsize{$\pm$3.38} \\                    
                                   
                                   \hline\hline
\multirow{6}{*}{\textbf{Pubmed}}   & GCN                & 61.3\scriptsize{$\pm$1.52}                    & 50.2\scriptsize{$\pm$2.08}  & 44.3\scriptsize{$\pm$1.43}          \\ 
                                  & GCNJaccard                  & 62.7\scriptsize{$\pm$1.25}                 & 51.9\scriptsize{$\pm$1.53}    &  45.1\scriptsize{$\pm$2.04}    \\ 
                                  & RGCN                     & 58.4\scriptsize{$\pm$1.74}                   & 49.0\scriptsize{$\pm$1.65}    &  43.9\scriptsize{$\pm$1.29}    \\
                                   & GAT            & 62.7\scriptsize{$\pm$1.68}           & 50.2\scriptsize{$\pm$2.35}          & 43.7\scriptsize{$\pm$2.43}          \\ \cline{2-5} 
                                  & AIB-Cat  & 64.5\scriptsize{$\pm$2.13} & 50.9\scriptsize{$\pm$3.83}          & 43.0\scriptsize{$\pm$3.73}          \\
                                   & AIB-Bern  & 61.1\scriptsize{$\pm$2.70} & 47.8\scriptsize{$\pm$3.65}          & 42.0\scriptsize{$\pm$4.21}          \\
                           
                                   & \textbf{GIB-Cat}          & \textbf{67.1}\scriptsize{$\pm$4.33}           & \textbf{57.2}\scriptsize{$\pm$5.27} & \textbf{51.5}\scriptsize{$\pm$4.84} \\
                            & \textbf{GIB-Bern}          & \textbf{64.9}\scriptsize{$\pm$2.52}           & \textbf{54.7}\scriptsize{$\pm$1.83} & \textbf{48.2}\scriptsize{$\pm$2.10} \\         
                                   
                                   \hline\hline
\multirow{6}{*}{\textbf{Citeseer}} & GCN                     & \textbf{55.9}\scriptsize{$\pm$1.33}           & 40.6\scriptsize{$\pm$1.83}          & 32.8\scriptsize{$\pm$2.19}          \\ 
                                  & GCNJaccard         & \textbf{56.8}\scriptsize{$\pm$1.49}               & 41.3\scriptsize{$\pm$1.81} &    33.1\scriptsize{$\pm$2.27}      \\ 
                                  & RGCN                & 51.4\scriptsize{$\pm$2.00}                  & 36.5\scriptsize{$\pm$2.38}     & 29.5\scriptsize{$\pm$2.17}    \\
                                   & GAT                   & 55.8\scriptsize{$\pm$1.43}           & 40.8\scriptsize{$\pm$1.77}          & 33.8\scriptsize{$\pm$1.93}          \\ \cline{2-5}
                                  & AIB-Cat        & 55.1\scriptsize{$\pm$1.26}           & 43.1\scriptsize{$\pm$2.46} & 35.6\scriptsize{$\pm$3.19}          \\

                                   & \textbf{AIB-Bern}           & 55.8\scriptsize{$\pm$2.01}  & \textbf{43.3}\scriptsize{$\pm$1.67}          & \textbf{36.3}\scriptsize{$\pm$2.47} \\ 
                                   
                      & GIB-Cat        & 54.9\scriptsize{$\pm$1.39}           & 42.0\scriptsize{$\pm$1.92} & 34.8\scriptsize{$\pm$1.75}          \\  
                                   
                          & \textbf{GIB-Bern}           & 54.4\scriptsize{$\pm$5.98}  & \textbf{50.3}\scriptsize{$\pm$4.33}          & \textbf{46.1}\scriptsize{$\pm$2.47} \\             
                                   \hline
\end{tabular}
}
\end{center}
\vskip -0.05in
\vspace{-0.2cm}
\end{table}

\subsection{Only Feature Attacks}
\label{sec:feature_attacks}
To further check the effectiveness of IB for node features, we inject random perturbation into the node features. Specifically, after the models are trained, we add independent Gaussian noise to each dimension of the node features for all nodes with increasing amplitude. Specifically, we use the mean of the maximum value of each node's feature as the reference amplitude $r$, and for each feature dimension of each node we add Gaussian noise $\lambda\cdot r \cdot\epsilon$, where $\epsilon\sim N(0,1)$, and $\lambda$ is the feature noise ratio. We test the models' performance with $\lambda\in\{$0.5, 1, 1.5$\}$. Table \ref{table:feature_noise} shows the results. We see across different feature noise ratios, both \GIBCat{} and \GIBBern{} consistently outperforms other models without IB, especially when the feature noise ratio is large ($\lambda=1.5$), and the AIB models with only structure IB performs slightly worse or equivalent to the GIB models. This shows that GIB makes the model more robust when the feature attack becomes the main source of perturbation.

\vspace{-0.15cm}
\section{Conclusion and Discussion}
\vspace{-0.1cm}

In this work, we have introduced Graph Information Bottleneck (GIB), an information-theoretic principle for learning representations that capture minimal sufficient information from graph-structured data. We have also demonstrated the efficacy of GIB by evaluating the robustness of the GAT model trained under the GIB principle on adversarial attacks. Our general framework leaves many interesting questions for future investigation. For example, are there any other better instantiations of GIB, especially in capturing discrete structural information? If incorporated with a node for global aggregation, can GIB break the limitation of the local-dependence assumption? May GIB be applied to other graph-related tasks including link prediction and graph classification?


\section*{Broader Impact}

\textbf{Who may benefit from this research:} Graphs have been used to represent a vast amount of real-world data from social science \cite{cho2011friendship}, biology \cite{mason2007graph}, geographical mapping \cite{barthelemy2011spatial}, finances \cite{kaastra1996designing} and recommender systems \cite{ying2018graph}, because of their flexibility in modeling both the relation among the data (structures) and the content of the data (features). Graph neural networks (GNN), naturally entangle both aspects of the data in the most expressive way, have attracted unprecedented attention from both academia and industry across a wide range of disciplines. However, GNNs share a common issue with other techniques based on neural networks. They are very sensitive to noise of data and are fragile to model attacks. This drawback yields the potential safety problems to deploy GNNs in the practical systems or use them to process data in those disciplines that heavily emphasize unbiased analysis. The Graph Information Bottleneck (GIB) principle proposed in this work paves a principled way to alleviate the above problem by increasing the robustness of GNN models. Our work further releases the worries about the usage of GNN techniques in practical systems, such as recommender systems, social media, or to analyze data for other disciplines, including physics, biology, social science. Ultimately, our work increases the interaction between AI, machine learning techniques and other aspects of our society, and could achieve far-reaching impact.

\textbf{Who may be put at disadvantage from this research:} Not applicable.

\textbf{What are the consequences of failure of the system:} Not applicable.

\textbf{Does the task/method leverage biases in the data:} The proposed GIB principle and the GIB-GAT model as an instantiation of GIB leverage the node features and structural information which in general are not believed to include undesirable biases. The datasets to evaluate our approaches are among the most widely-used benchmarks, which in general are not believed to include undesirable biases as well.

\begin{ack}
We thank the anonymous reviewers for providing feedback on our manuscript.
Hongyu Ren is supported by the Masason Foundation Fellowship. 
Jure Leskovec is a Chan Zuckerberg Biohub investigator.
We also gratefully acknowledge the support of
DARPA under Nos. FA865018C7880 (ASED), N660011924033 (MCS);
ARO under Nos. W911NF-16-1-0342 (MURI), W911NF-16-1-0171 (DURIP);
NSF under Nos. OAC-1835598 (CINES), OAC-1934578 (HDR), CCF-1918940 (Expeditions), IIS-2030477 (RAPID);
Stanford Data Science Initiative, 
Wu Tsai Neurosciences Institute,
Chan Zuckerberg Biohub,
Amazon, Boeing, JPMorgan Chase, Docomo, Hitachi, JD.com, KDDI, NVIDIA, Dell. 
\end{ack}

\hidecomments{\jure{Make sure to check the references. The GCN paper is cited twice. Make sure. that conference names are always spelled the same way}}

\bibliography{reference}

\begin{thebibliography}{10}
\providecommand{\url}[1]{#1}
\csname url@samestyle\endcsname
\providecommand{\newblock}{\relax}
\providecommand{\bibinfo}[2]{#2}
\providecommand{\BIBentrySTDinterwordspacing}{\spaceskip=0pt\relax}
\providecommand{\BIBentryALTinterwordstretchfactor}{4}
\providecommand{\BIBentryALTinterwordspacing}{\spaceskip=\fontdimen2\font plus
\BIBentryALTinterwordstretchfactor\fontdimen3\font minus
  \fontdimen4\font\relax}
\providecommand{\BIBforeignlanguage}[2]{{%
\expandafter\ifx\csname l@#1\endcsname\relax
\typeout{** WARNING: IEEEtran.bst: No hyphenation pattern has been}%
\typeout{** loaded for the language `#1'. Using the pattern for}%
\typeout{** the default language instead.}%
\else
\language=\csname l@#1\endcsname
\fi
#2}}
\providecommand{\BIBdecl}{\relax}
\BIBdecl

\bibitem{hamilton2017inductive}
W.~Hamilton, Z.~Ying, and J.~Leskovec, ``Inductive representation learning on
  large graphs,'' in \emph{Advances in neural information processing systems},
  2017.

\bibitem{kipf2016variational}
T.~N. Kipf and M.~Welling, ``Variational graph auto-encoders,'' \emph{arXiv
  preprint arXiv:1611.07308}, 2016.

\bibitem{kipf2017semi}
------, ``Semi-supervised classification with graph convolutional networks,''
  in \emph{International Conference on Learning Representations}, 2017.

\bibitem{li2019optimizing}
P.~Li, I.~Chien, and O.~Milenkovic, ``Optimizing generalized pagerank methods
  for seed-expansion community detection,'' in \emph{Advances in Neural
  Information Processing Systems}, 2019.

\bibitem{velickovic2018graph}
P.~Veličković, G.~Cucurull, A.~Casanova, A.~Romero, P.~Liò, and Y.~Bengio,
  ``Graph attention networks,'' in \emph{International Conference on Learning
  Representations}, 2018.

\bibitem{chen2018fastgcn}
J.~Chen, T.~Ma, and C.~Xiao, ``Fast{GCN}: Fast learning with graph
  convolutional networks via importance sampling,'' in \emph{International
  Conference on Learning Representations}, 2018.

\bibitem{klicpera_predict_2019}
J.~Klicpera, A.~Bojchevski, and S.~G{\"u}nnemann, ``Predict then propagate:
  Graph neural networks meet personalized pagerank,'' in \emph{International
  Conference on Learning Representations}, 2019.

\bibitem{xu2018how}
K.~Xu, W.~Hu, J.~Leskovec, and S.~Jegelka, ``How powerful are graph neural
  networks?'' in \emph{International Conference on Learning Representations},
  2019.

\bibitem{pmlr-v97-you19b}
J.~You, R.~Ying, and J.~Leskovec, ``Position-aware graph neural networks,'' in
  \emph{International Conference on Machine Learning}, 2019.

\bibitem{Pei2020Geom-GCN}
H.~Pei, B.~Wei, K.~C.-C. Chang, Y.~Lei, and B.~Yang, ``Geom-gcn: Geometric
  graph convolutional networks,'' in \emph{International Conference on Learning
  Representations}, 2020.

\bibitem{maron2019provably}
H.~Maron, H.~Ben-Hamu, H.~Serviansky, and Y.~Lipman, ``Provably powerful graph
  networks,'' in \emph{Advances in Neural Information Processing Systems},
  2019.

\bibitem{murphy2019relational}
R.~Murphy, B.~Srinivasan, V.~Rao, and B.~Riberio, ``Relational pooling for
  graph representations,'' in \emph{International Conference on Machine
  Learning}, 2019.

\bibitem{chen2019equivalence}
Z.~Chen, S.~Villar, L.~Chen, and J.~Bruna, ``On the equivalence between graph
  isomorphism testing and function approximation with gnns,'' in \emph{Advances
  in Neural Information Processing Systems}, 2019.

\bibitem{Hou2020Measuring}
Y.~Hou, J.~Zhang, J.~Cheng, K.~Ma, R.~T.~B. Ma, H.~Chen, and M.-C. Yang,
  ``Measuring and improving the use of graph information in graph neural
  networks,'' in \emph{International Conference on Learning Representations},
  2020.

\bibitem{zugner2018adversarial}
D.~Z{\"u}gner, A.~Akbarnejad, and S.~G{\"u}nnemann, ``Adversarial attacks on
  neural networks for graph data,'' in \emph{Proceedings of the 24th ACM SIGKDD
  International Conference on Knowledge Discovery \& Data Mining}, 2018.

\bibitem{dai2018adversarial}
H.~Dai, H.~Li, T.~Tian, X.~Huang, L.~Wang, J.~Zhu, and L.~Song, ``Adversarial
  attack on graph structured data,'' \emph{arXiv preprint arXiv:1806.02371},
  2018.

\bibitem{cover2012elements}
T.~M. Cover and J.~A. Thomas, \emph{Elements of information theory}.\hskip 1em
  plus 0.5em minus 0.4em\relax John Wiley \& Sons, 2012.

\bibitem{tishby2000information}
N.~Tishby, F.~C. Pereira, and W.~Bialek, ``The information bottleneck method,''
  \emph{arXiv preprint physics/0004057}, 2000.

\bibitem{tishby2015deep}
N.~Tishby and N.~Zaslavsky, ``Deep learning and the information bottleneck
  principle,'' in \emph{2015 IEEE Information Theory Workshop (ITW)}.\hskip 1em
  plus 0.5em minus 0.4em\relax IEEE, 2015.

\bibitem{dirac1981principles}
P.~A.~M. Dirac, \emph{The principles of quantum mechanics}.\hskip 1em plus
  0.5em minus 0.4em\relax Oxford university press, 1981, no.~27.

\bibitem{alemi2016deep}
A.~A. Alemi, I.~Fischer, J.~V. Dillon, and K.~Murphy, ``Deep variational
  information bottleneck,'' \emph{arXiv preprint arXiv:1612.00410}, 2016.

\bibitem{poole2019variational}
B.~Poole, S.~Ozair, A.~Van Den~Oord, A.~Alemi, and G.~Tucker, ``On variational
  bounds of mutual information,'' in \emph{International Conference on Machine
  Learning}, 2019.

\bibitem{nguyen2010estimating}
X.~Nguyen, M.~J. Wainwright, and M.~I. Jordan, ``Estimating divergence
  functionals and the likelihood ratio by convex risk minimization,''
  \emph{IEEE Transactions on Information Theory}, 2010.

\bibitem{jang2016categorical}
E.~Jang, S.~Gu, and B.~Poole, ``Categorical reparameterization with
  gumbel-softmax,'' in \emph{International Conference on Learning
  Representations}, 2017.

\bibitem{maddison2016concrete}
C.~J. Maddison, A.~Mnih, and Y.~W. Teh, ``The concrete distribution: A
  continuous relaxation of discrete random variables,'' in \emph{International
  Conference on Learning Representations}, 2017.

\bibitem{fischer2020ceb}
I.~Fischer and A.~A. Alemi, ``Ceb improves model robustness,'' \emph{arXiv
  preprint arXiv:2002.05380}, 2020.

\bibitem{dilokthanakul2016deep}
N.~Dilokthanakul, P.~A. Mediano, M.~Garnelo, M.~C. Lee, H.~Salimbeni,
  K.~Arulkumaran, and M.~Shanahan, ``Deep unsupervised clustering with gaussian
  mixture variational autoencoders,'' \emph{arXiv preprint arXiv:1611.02648},
  2016.

\bibitem{oord2018representation}
A.~v.~d. Oord, Y.~Li, and O.~Vinyals, ``Representation learning with
  contrastive predictive coding,'' \emph{arXiv preprint arXiv:1807.03748},
  2018.

\bibitem{gilmer2017neural}
J.~Gilmer, S.~S. Schoenholz, P.~F. Riley, O.~Vinyals, and G.~E. Dahl, ``Neural
  message passing for quantum chemistry,'' in \emph{Proceedings of the 34th
  International Conference on Machine Learning-Volume 70}.\hskip 1em plus 0.5em
  minus 0.4em\relax JMLR. org, 2017.

\bibitem{li2018adaptive}
R.~Li, S.~Wang, F.~Zhu, and J.~Huang, ``Adaptive graph convolutional neural
  networks,'' in \emph{Thirty-second AAAI conference on artificial
  intelligence}, 2018.

\bibitem{xu2018representation}
K.~Xu, C.~Li, Y.~Tian, T.~Sonobe, K.-i. Kawarabayashi, and S.~Jegelka,
  ``Representation learning on graphs with jumping knowledge networks,''
  \emph{arXiv preprint arXiv:1806.03536}, 2018.

\bibitem{zhang2018gaan}
J.~Zhang, X.~Shi, J.~Xie, H.~Ma, I.~King, and D.-Y. Yeung, ``Gaan: Gated
  attention networks for learning on large and spatiotemporal graphs,''
  \emph{arXiv preprint arXiv:1803.07294}, 2018.

\bibitem{zhu2019robust}
D.~Zhu, Z.~Zhang, P.~Cui, and W.~Zhu, ``Robust graph convolutional networks
  against adversarial attacks,'' in \emph{Proceedings of the 25th ACM SIGKDD
  International Conference on Knowledge Discovery \& Data Mining}, 2019.

\bibitem{wu2019adversarial}
H.~Wu, C.~Wang, Y.~Tyshetskiy, A.~Docherty, K.~Lu, and L.~Zhu, ``Adversarial
  examples for graph data: Deep insights into attack and defense,'' in
  \emph{International Joint Conference on Artificial Intelligence, IJCAI},
  2019.

\bibitem{entezari2020all}
N.~Entezari, S.~A. Al-Sayouri, A.~Darvishzadeh, and E.~E. Papalexakis, ``All
  you need is low (rank) defending against adversarial attacks on graphs,'' in
  \emph{Proceedings of the 13th International Conference on Web Search and Data
  Mining}, 2020.

\bibitem{zugner2019certifiable}
D.~Z{\"u}gner and S.~G{\"u}nnemann, ``Certifiable robustness and robust
  training for graph convolutional networks,'' in \emph{Proceedings of the 25th
  ACM SIGKDD International Conference on Knowledge Discovery \& Data Mining},
  2019.

\bibitem{dgi}
P.~Veli{\v{c}}kovi{\'c}, W.~Fedus, W.~L. Hamilton, P.~Li{\`o}, Y.~Bengio, and
  R.~D. Hjelm, ``Deep graph infomax,'' \emph{arXiv preprint arXiv:1809.10341},
  2018.

\bibitem{gmi}
Z.~Peng, W.~Huang, M.~Luo, Q.~Zheng, Y.~Rong, T.~Xu, and J.~Huang, ``Graph
  representation learning via graphical mutual information maximization,'' in
  \emph{Proceedings of The Web Conference 2020}, 2020.

\bibitem{sun2019infograph}
F.-Y. Sun, J.~Hoffmann, and J.~Tang, ``Infograph: Unsupervised and
  semi-supervised graph-level representation learning via mutual information
  maximization,'' \emph{arXiv preprint arXiv:1908.01000}, 2019.

\bibitem{vdb}
X.~B. Peng, A.~Kanazawa, S.~Toyer, P.~Abbeel, and S.~Levine, ``Variational
  discriminator bottleneck: Improving imitation learning, inverse rl, and gans
  by constraining information flow,'' \emph{arXiv preprint arXiv:1810.00821},
  2018.

\bibitem{betavae}
I.~Higgins, L.~Matthey, A.~Pal, C.~Burgess, X.~Glorot, M.~Botvinick,
  S.~Mohamed, and A.~Lerchner, ``beta-vae: Learning basic visual concepts with
  a constrained variational framework.'' in \emph{International Conference on
  Learning Representations}, 2017.

\bibitem{hjelm2018learning}
R.~D. Hjelm, A.~Fedorov, S.~Lavoie-Marchildon, K.~Grewal, P.~Bachman,
  A.~Trischler, and Y.~Bengio, ``Learning deep representations by mutual
  information estimation and maximization,'' in \emph{International Conference
  on Learning Representations}, 2019.

\bibitem{sen2008collective}
P.~Sen, G.~Namata, M.~Bilgic, L.~Getoor, B.~Galligher, and T.~Eliassi-Rad,
  ``Collective classification in network data,'' \emph{AI magazine}, 2008.

\bibitem{cho2011friendship}
E.~Cho, S.~A. Myers, and J.~Leskovec, ``Friendship and mobility: user movement
  in location-based social networks,'' in \emph{Proceedings of the 17th ACM
  SIGKDD international conference on Knowledge discovery and data mining},
  2011.

\bibitem{mason2007graph}
O.~Mason and M.~Verwoerd, ``Graph theory and networks in biology,'' \emph{IET
  systems biology}, 2007.

\bibitem{barthelemy2011spatial}
M.~Barth{\'e}lemy, ``Spatial networks,'' \emph{Physics Reports}, 2011.

\bibitem{kaastra1996designing}
I.~Kaastra and M.~Boyd, ``Designing a neural network for forecasting
  financial,'' \emph{Neurocomputing}, 1996.

\bibitem{ying2018graph}
R.~Ying, R.~He, K.~Chen, P.~Eksombatchai, W.~L. Hamilton, and J.~Leskovec,
  ``Graph convolutional neural networks for web-scale recommender systems,'' in
  \emph{Proceedings of the 24th ACM SIGKDD International Conference on
  Knowledge Discovery \& Data Mining}, 2018.

\bibitem{fischer2020conditional}
I.~Fischer, ``The conditional entropy bottleneck,'' \emph{arXiv preprint
  arXiv:2002.05379}, 2020.

\bibitem{NEURIPS2019_9015}
A.~Paszke, S.~Gross, F.~Massa, A.~Lerer, J.~Bradbury, G.~Chanan, T.~Killeen,
  Z.~Lin, N.~Gimelshein, L.~Antiga, A.~Desmaison, A.~Kopf, E.~Yang, Z.~DeVito,
  M.~Raison, A.~Tejani, S.~Chilamkurthy, B.~Steiner, L.~Fang, J.~Bai, and
  S.~Chintala, ``Pytorch: An imperative style, high-performance deep learning
  library,'' in \emph{Advances in Neural Information Processing Systems 32},
  H.~Wallach, H.~Larochelle, A.~Beygelzimer, F.~d\textquotesingle
  Alch\'{e}-Buc, E.~Fox, and R.~Garnett, Eds.\hskip 1em plus 0.5em minus
  0.4em\relax Curran Associates, Inc., 2019.

\bibitem{FeyPyG}
M.~Fey and J.~E. Lenssen, ``Fast graph representation learning with {PyTorch
  Geometric},'' in \emph{ICLR Workshop on Representation Learning on Graphs and
  Manifolds}, 2019.

\end{thebibliography}
\bibliographystyle{IEEEtran}

\newpage
 
\appendix

\begin{center}
\begin{huge}
\textbf{Appendix}
\end{huge}
\end{center}

\section{Preliminaries for Information Bottleneck}

Here we briefly review the Information Bottleneck (IB) principle and its application to representation learning.

Given the input data $\D$ and target $Y$, and an stochastic encoding $Z$ of $\D$ by $\mathbb{P}(Z|\D)$ that satisfies the Markov chain $Z-\D-Y$, IB has the following objective:
\begin{equation}
\label{eq:IB}
\min_{\mathbb{P}(Z|\D)}{\IB_\beta(\mathcal{D},Y;Z)}:=\left[ - I(Y;Z)+\beta I(\mathcal{D};Z) \right]
\end{equation}

It also has an equivalent form:
\begin{equation}
\label{eq:IB_constraint_opt}
\max_{\mathbb{P}(Z|\D):I(\D;Z)\le I_c}{I(Y;Z)}
\end{equation}

Intuitively, Eq. \eqref{eq:IB} or \eqref{eq:IB_constraint_opt} encourages the representation $Z$ to maximally capture the information in $Y$, while controlling the complexity of the representation in terms of $I(\D;Z)$. When increasing $\beta$ from 0 to some large value, we are essentially using a straight line with slope $\beta$ to sweep out the Pareto frontier of $I(Y;Z)$ vs. $I(X;Z)$ as given by Eq. \eqref{eq:IB_constraint_opt}.

\begin{figure}[h]
\begin{center}
\includegraphics[width=0.6\columnwidth]{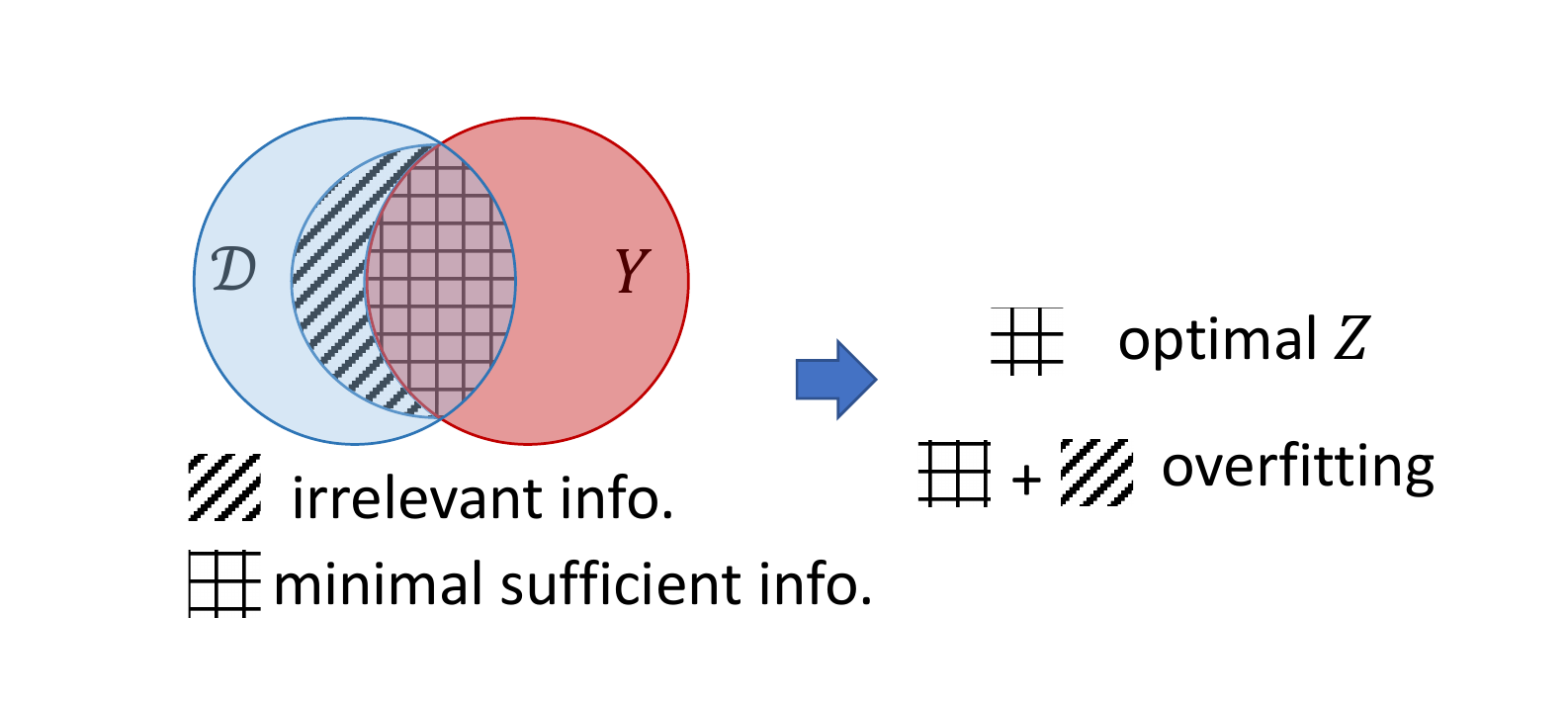}
\end{center}
\caption{
Information diagram for the Information Bottleneck (IB). Also plotted are the minimal sufficient information as covered by $I(\D;Y)$ and overfitting part that occupies parts of $H(\D|Y)$.}
\label{fig:IB_venn}
\end{figure}

Using the information diagram (Fig. \ref{fig:IB_venn}), where we represent the information of $\D$, $Y$ as circles and their shared part as the overlapping region of the circles, then IB encourages $Z$ to cover as much of the $I(\D;Y)$ as possible, and cover as little of $H(\D|Y)$ (the irrelevant information part) as possible. An optimal representation is defined as the minimal sufficient representation \cite{fischer2020conditional} that only covers $I(\D;Y)$. In practice, due to the expressiveness of the models and different choices of $\beta$ in Eq. \eqref{eq:IB}, this optimal information can hardly be reached, and may only be approached. It is an interesting future direction to study that when sweeping $\beta$, how near it is to the optimal representation on the diagram of $I(Y;Z)$ vs. $I(X;Z)$.

\section{Proof for Proposition~\ref{prop:IYZ}}
\label{app:proof3_1}
We restate Proposition~\ref{prop:IXZ}:  For any PDFs $\mathbb{Q}_1(Y_v|Z_{X,v}^{(L)})$ for $v\in V$ and $\mathbb{Q}_2(Y)$, we have 
\begin{align}\label{eq:iyz_app}
I(Y;Z_X^{(L)}) \geq 1 +  \mathbb{E}\left[\log\frac{\prod_{v\in V}\mathbb{Q}_1(Y_v|Z_{X,v}^{(L)})}{\mathbb{Q}_2(Y)}\right]  +\mathbb{E}_{\mathbb{P}(Y)\mathbb{P}(Z_{X}^{(L)})}\left[\frac{\prod_{v\in V}\mathbb{Q}_1(Y_v|Z_{X,v}^{(L)})}{\mathbb{Q}_2(Y)}\right] 
\end{align}
\begin{proof}
We use the Nguyen, Wainright \& Jordan's bound $I_{\text{NWJ}}$~\cite{poole2019variational,nguyen2010estimating}:
\begin{lemma}\cite{poole2019variational,nguyen2010estimating}
For any two random variables $X_1, X_2$ and any function $g: g(X_1, X_2)\in \mathbb{R}$, we have 
\begin{align*}
    I(X_1, X_2) \geq \mathbb{E}\left[g(X_1, X_2)\right] - \mathbb{E}_{\mathbb{P}(X_1)\mathbb{P}(X_2)}\left[\exp(g(X_1, X_2)-1)\right].
\end{align*}
\end{lemma}
We use the above lemma to $(Y, Z_X^{(L)})$ and plug in $g(Y, Z_X^{(L)}) = 1 + \log\frac{\prod_{v\in V}\mathbb{Q}_1(Y_v|Z_{X,v}^{(L)})}{\mathbb{Q}_2(Y)}$. 
\end{proof}

\section{Proof for Proposition~\ref{prop:IXZ}}
\label{app:proof3_2}
We restate Proposition~\ref{prop:IXZ}: For any groups of indices $S_X, S_A\subset [L]$ such that $\mathcal{D} \perp Z_X^{(L)} | \{Z_X^{(l)}\}_{l\in S_X}\cup \{Z_A^{(l)}\}_{l\in S_A}$, and for any probabilistic distributions $\mathbb{Q}(Z_X^{(l)})$, $l\in S_X$, and $\mathbb{Q}(Z_A^{(l)})$, $l\in S_A$,
\begin{align}\label{eq:ixz_app}
&I(\mathcal{D}; Z_X^{(L)}) \leq I(\mathcal{D}; \{Z_X^{(l)}\}_{l\in S_X}\cup \{Z_A^{(l)}\}_{l\in S_A}) \leq \sum_{l\in S_X}\text{XIB}^{(l)} + \sum_{l\in S_A}\text{AIB}^{(l)}, \text{where}  \\
&\text{AIB}^{(l)} = \mathbb{E}\left[\log \frac{\mathbb{P}(Z_A^{(l)}|A, Z_X^{(l-1)})}{\mathbb{Q}(Z_A^{(l)})} \right], \text{XIB}^{(l)} = \mathbb{E}\left[\log \frac{\mathbb{P}(Z_X^{(l)}|Z_X^{(l-1)}, Z_A^{(l)})}{\mathbb{Q}(Z_X^{(l)})} \right],
\end{align}
\begin{proof}
The first inequality $I(\mathcal{D}; Z_X^{(L)}) \leq I(\mathcal{D}; \{Z_X^{(l)}\}_{l\in S_X}\cup \{Z_A^{(l)}\}_{l\in S_A})$ directly results from the data processing inequality~\cite{cover2012elements} and the Markov property $\mathcal{D} \perp Z_X^{(L)} | \{Z_X^{(l)}\}_{l\in S_X}\cup \{Z_A^{(l)}\}_{l\in S_A}$. 

To prove the second inequality, we define an order ``$\prec$'' of random variables in $\{Z_X^{(l)}\}_{l\in S_X}\cup \{Z_A^{(l)}\}_{l\in S_A}$ such that 1) for two different integers $l,\,l'$, $Z_X^{(l)}, Z_A^{(l)}\prec Z_X^{(l')}, Z_A^{(l')}$; 2) For one integer $l$, $Z_A^{(l)}\prec Z_X^{(l)}$. Based on the order, define a sequence of sets
\begin{align*}
    H_A^{(l)} & = \{Z_X^{(l_1)}, Z_A^{(l_2)}| l_1< l, l_2 < l, l_1\in S_X, l_2\in S_A \}, \\
     H_X^{(l)} & = \{Z_X^{(l_1)}, Z_A^{(l_2)}| l_1< l, l_2\leq l, l_1\in S_X, l_2\in S_A \}. \\
\end{align*}
We may decompose $I(\mathcal{D}; \{Z_X^{(l)}\}_{l\in S_X}\cup \{Z_A^{(l)}\}_{l\in S_A})$ with respect to this order
\begin{align*}
    I(\mathcal{D}; \{Z_X^{(l)}\}_{l\in S_X}\cup \{Z_A^{(l)}\}_{l\in S_A}) = \sum_{l\in S_A} I(\mathcal{D}; Z_A^{(l)} | H_A^{(l)}) + \sum_{l\in S_X} I(\mathcal{D}; Z_X^{(l)} | H_X^{(l)}).
\end{align*}
Next, we bound each term in the RHS
\begin{align*}
I(\mathcal{D}; Z_A^{(l)} | H_A^{(l)}) &\stackrel{\text{1)}}{\leq} I(\mathcal{D}, Z_X^{(l-1)}; Z_A^{(l)} | H_A^{(l)}) \\
&\stackrel{\text{2)}}{=} I(Z_X^{(l-1)}, A ; Z_A^{(l)} | H_A^{(l)}) + I(X; Z_A^{(l)} | H_A^{(l)}, A, Z_X^{(l-1)}) \\
&\stackrel{\text{3)}}{=} I(Z_X^{(l-1)}, A ; Z_A^{(l)} | H_A^{(l)}) + 0\\
&\stackrel{\text{4)}}{\leq} I(Z_X^{(l-1)}, A ; Z_A^{(l)}) \\
&\stackrel{\text{5)}}{=} \text{AIB}^{(l)} - \text{KL}(\mathbb{P}(Z_A^{(l)})|| \mathbb{Q}(Z_A^{(l)})) \leq \text{AIB}^{(l)} \\
I(\mathcal{D}; Z_X^{(l)} | H_X^{(l)}) &\stackrel{\text{1)}}{\leq} I(\mathcal{D}, Z_X^{(l-1)}, Z_A^{(l)}; Z_X^{(l)} | H_X^{(l)}) \\
&\stackrel{\text{2)}}{=} I(Z_X^{(l-1)}, Z_A^{(l)} ; Z_X^{(l)} | H_X^{(l)}) + I(\mathcal{D}; Z_X^{(l)} | H_X^{(l)}, Z_X^{(l-1)}, Z_A^{(l)}) \\
&\stackrel{\text{3)}}{=} I(Z_X^{(l-1)}, Z_A^{(l)} ; Z_X^{(l)} | H_X^{(l)}) + 0 \\
&\stackrel{\text{4)}}{\leq} I(Z_X^{(l-1)}, Z_A^{(l)} ; Z_X^{(l)}) \\
&\stackrel{\text{5)}}{=} \text{XIB}^{(l)} - \text{KL}(\mathbb{P}(Z_X^{(l)})|| \mathbb{Q}(Z_X^{(l)})) \leq \text{XIB}^{(l)}
\end{align*}
where $1), 2)$ use the basic properties of mutual information, $3)$ uses $X \perp Z_A^{(l)} | \{A, Z_X^{(l-1)}\}$ and $\mathcal{D} \perp Z_X^{(l)} | \{Z_A^{(l-1)}, Z_X^{(l-1)}\}$, $4)$ uses $H_A^{(l)} \perp Z_A^{(l)} | \{Z_X^{(l-1)}, A\}$ and  $H_X^{(l)} \perp Z_X^{(l)} | \{Z_X^{(l-1)}, Z_A^{(l)}\}$ and 5) uses the definitions of $\text{AIB}^{(l)}$ and $\text{XIB}^{(l)}$.   
\end{proof}

\section{The Contrastive Loss Derived from the Variational Bound Eq. (2)}
\label{app:cons-loss}
To characterize Eq.~\eqref{eq:iyz}, We may also use a contrastive loss~\cite{oord2018representation,poole2019variational} which empirically may sometimes improve the robustness of the model. Concretely, we keep $\mathbb{Q}_1(Y_v|Z_{X,v}^{(L)})$ as the same as that to derive Eq.~\eqref{eq:prac-iyz}, i.e., $\mathbb{Q}_1(Y_v|Z_{X,v}^{(L)}) = \text{Cat}(Z_{X,v}^{(L)}W_{\text{out}})$ and set $\mathbb{Q}_2(Y) = \mathbb{E}_{\mathbb{P}(Z_{X}^{(L)})\mathbb{P}(Z_{X}^{'(L)})}[\prod_{v\in V} \frac{1}{2}(\mathbb{Q}_1(Y_v|Z_{X,v}^{(L)})+\mathbb{Q}_1(Y_v|Z_{X,v}^{'(L)}))]$. Here, $\mathbb{P}(Z_{X}^{'(L)})$ refers to the distribution of the last-layer node representation after we replace $A$ with a random graph structure $A'\in \mathbb{R}^{n \times n}$ where $A'$ is uniformly sampled with the constraint that $A'$ has the same number of edges as $A$. When using this contrastive loss, we simply use the estimation of $\mathbb{Q}_2(Y)$ based on the sampled $Z_{X,v}^{(L)}$ and $Z_{X,v}^{'(L)}$. Moreover, the last term of Eq.~\eqref{eq:iyz} is empirically closed to 1 and thus we ignore it and other constants in Eq. \eqref{eq:iyz}. Overall, we have the substitution for the contrastive loss,
\begin{align}\label{eq:prac-iyz_app}
    I(Y;Z_X^{(L)}) \rightarrow \sum_{v\in V}\left[\log(h(Y_v;Z_{X,v}^{(L)})) - \log(h(Y_v;Z_{X,v}^{(L)}) + h(Y_v;Z_{X,v}^{'(L)})) \right],
\end{align}
where $h(Y_v;Z_{X,v}) = \frac{\exp(Z_{X,v}W_{\text{out}}[Y_v])}{\sum_{i=1}^K\exp(Z_{X,v}W_{\text{out}}[i])}$.

\section{Permutation Invariance of \GIBCat{} and \GIBBern{}} \label{app:perm-inv}
Let $\Pi\in\mathbb{R}^{n\times n}$ denote a permutation matrix where each row and each column contains exactly one single 1 and the rest components are all 0's. For any variable in \GIBCat{} or \GIBBern{}, we use subscript $\Pi$ to denote the corresponding obtained variable after we permutate the node indices of the input data, i.e., $\mathcal{D}=(X, A) \rightarrow \Pi(\mathcal{D}) = (\Pi X, \Pi A \Pi^T)$. For example, $Z_{X,\Pi}^{(l)}$ denotes the node representations after $l$ layers of \GIBCat{} or \GIBBern{} based on the input data $\Pi(\mathcal{D})$. Moreover, the matrix $\Pi$ also defines a bijective mapping $\pi: V\rightarrow V$, where $\pi(v)=u$ iff $\Pi_{uv}=1$. We also use ``$\stackrel{d}{=}$'' to denote that two random variables share the same distribution.

Now, we formally restate the permutation invariant property of \GIBCat{} and \GIBBern{}: Suppose $\Pi\in\mathbb{R}^{n\times n}$ is any permutation matrix, if the input graph-structured data becomes $\Pi(\mathcal{D}) = (\Pi X, \Pi A \Pi^T)$, the corresponding node representations output by \GIBCat{} or \GIBBern{} satisfy $Z_{X,\Pi}^{(L)}\stackrel{d}{=} \Pi Z_{X}^{(L)}$ where $Z_{X}^{(L)}$ is the output node representations based on the original input data $\mathcal{D} = (X,  A )$. 
\begin{proof}
We use induction to prove this result. Specifically, we only need to show that for a certain $l\in [L]$, if node representations $Z_{X,\Pi}^{(l-1)}\stackrel{d}{=} \Pi Z_{X}^{(l-1)}$ and $A \rightarrow \Pi A \Pi^T$, then the refined node representations $Z_{X,\Pi}^{(l)}\stackrel{d}{=}\Pi Z_{X}^{(l)}$. To prove this statement, we go through Algorithm 1 step by step. 
\begin{itemize}
    \item Step 2 implies $\tilde{Z}_{X,v,\Pi}^{(l-1)} \stackrel{d}{=} \tilde{Z}_{X,\pi(v)}^{(l-1)} $ because $\tau$ is element-wise operation.
    \item Steps 3: For both NeighborSample (categorical and Bernoulli) by Algorithm 2/3, the substeps 1-2 imply $\phi_{vt,\Pi}^{(l)} \stackrel{d}{=} \Pi\phi_{\pi(v)t}^{(l)}$. Here, we use $A \rightarrow \Pi A \Pi^T$ and thus $V_{vt}\rightarrow V_{\pi(v)t}$, and assume that $\phi_{vt,\Pi}^{(l)}, \phi_{\pi(v)t}^{(l)}$ are represented as vectors in $\mathbb{R}^{n\times 1}$ where their $u$th components, $\phi_{vt,\Pi,u}^{(l)}, \phi_{\pi(v)t, u}^{(l)}$, are $0$'s if $\pi^{-1}(u)\notin V_{vt}$. Substep 3, implies $Z_{A,v,\Pi}^{(l)} \stackrel{d}{=} \pi(Z_{A,\pi(v)}^{(l)})$ where $\pi(S)= \{\pi(v)|v\in S\}$ for some set $S\subseteq V$.
    \item Step 4 implies $\bar{Z}_{X,v,\Pi}^{(l)}  \stackrel{d}{=} \bar{Z}_{X,\pi(v)}^{(l)}$.
    \item Steps 5-6 imply $\mu_{v,\Pi}^{(l)} \stackrel{d}{=} \mu_{\pi(v)}^{(l)}$, $\sigma_{v,\Pi}^{2(l)} \stackrel{d}{=} \sigma_{\pi(v)}^{2(l)}$.
    \item Step 7 implies $Z_{X,v, \Pi}^{(l)} \stackrel{d}{=} Z_{X,\pi(v)}^{(l)}$.
\end{itemize}
which indicates $Z_{X,\Pi}^{(l)} \stackrel{d}{=} \Pi Z_{X}^{(l)}$ and concludes the proof.
\end{proof}

\section{Summary of the Datasets}
Table \ref{table:datasets} summarizes statistics of the datasets (Cora, Pubmed, Citeseer \cite{sen2008collective}) we use, as well as the standard train-validation-test split we use in the experiments.
\label{app:datasets}
\begin{table}[ht]
\begin{center}
\caption{
Summary of the datasets and splits in our experiments.
}
\label{table:datasets}
\vskip 0.1in
\setlength{\tabcolsep}{4pt}  
\begin{tabular}{l | c c c }
\hline
\tiny
& \textbf{Cora} & \textbf{Pubmed} & \textbf{Citeseer}\\ [0.2ex]
\hline
\textbf{\# Nodes} & 2708  & 19717 & 3327 \\
\textbf{\# Edges} & 5429  & 44338 & 4732 \\
\textbf{\# Features/Node} & 1433 & 500 & 3703 \\
\textbf{\# Classes} & 7 & 3 & 6 \\
\textbf{\# Training Nodes} & 140 & 60 & 120 \\
\textbf{\# Validation Nodes} & 500 & 500 & 500 \\
\textbf{\# Test Nodes} & 1000 & 1000 & 1000 \\
\hline\noalign{\smallskip}

\end{tabular}
\end{center}
\end{table}

\section{Implementation Details for the \GIBCat{}, \GIBBern{} and Other Compared Models}
\label{app:implementation_models}

For all experiments and all models, the best models are selected according to the classification accuracy on the validation set. All models are trained with a total of 2000 epochs. For all experiments, we run it with 5 random seeds: 0, 1, 2, 3, 4 and report the average performance and standard deviation. The models are all trained on NVIDIA GeForce RTX 2080 GPUs, together with Intel(R) Xeon(R) Gold 6148 CPU @ 2.40GH CPUs. We use PyTorch \cite{NEURIPS2019_9015} and PyTorch Geometric \cite{FeyPyG} for constructing the GNNs and evaluation. Project website and code can be found at \url{http://snap.stanford.edu/gib/}. In Section \ref{app:implementation_GIB}, \ref{app:implementation_GCN_GAT} and \ref{app:implementation_defense_baseline}, we detail the hyperparameter setting for Section \ref{sec:robustness_adversary}, and in Section \ref{app:implementation_nettack} and \ref{sec:additinal_feature}, we provide additional details for the experiments.

\subsection{Implementation Details for the \GIBCat{} and \GIBBern{}}
\label{app:implementation_GIB}

The architecture of \GIBCat{} and \GIBBern{} follows Alg. 1 (and Alg. 2 and 3 for the respective neighbor-sampling). We follow GAT \cite{velickovic2018graph}'s default architecture, in which we use 8 attention heads, nonlinear activation $\tau$ as LeakyReLU, and feature dropout rate of 0.6 between layers. We follow GAT's default learning rate, i.e. 0.01 for Cora and Citeseer, and $5\times10^{-3}$ for Pubmed. As stated in the main text, the training objective is Eq. \eqref{eq:gib}, substituting in Eq. \eqref{eq:prac-ixz} and \eqref{eq:prac-iyz}. To allow more flexibility (in similar spirit as $\beta$-VAE \cite{betavae}), we allow the coefficient before $\widehat{\text{AIB}}$ and $\widehat{\text{XIB}}$ to be different, and denote them as $\beta_1$ and $\beta_2$. In summary, the objective is written as:

\begin{equation}
L=\sum_{v\in V}\text{Cross-Entropy}(Z_{X,v}^{(L)}W_{\text{out}}; Y_v)+\beta_1\sum_{l\in S_A}\widehat{\text{AIB}}^{(l)}+\beta_2 \sum_{l\in S_X}\widehat{\text{XIB}}^{(l)}
\end{equation}

In this work, we set the index set $S_A=[L]=\{1,2,...L\}$ and $S_X=\{L-1\}$, which satisfies Proposition \ref{prop:IXZ}. For $\widehat{\text{XIB}}$, we use mixture of Gaussians as the variational marginal distribution $\mathbb{Q}(Z_X)$. For the mixture of Gaussians, we use $m=100$ components with learnable weights, where each component is a diagonal Gaussian with learnable mean and standard deviation. This flexible variational marginal allows it to flexibly approximate the true marginal distribution $\mathbb{P}(Z_X)$. For the reparameterization in $\widehat{\text{AIB}}$,  we use Gumbel-softmax~\cite{jang2016categorical,maddison2016concrete} with temperature $\tau$. For \GIBCat{}, the number of neighbors $k$ to be sampled is a hyperparameter. For \GIBBern{}, we use Bernoulli($\alpha$) as the non-informative prior, where we fix $\alpha=0.5$. To facilitate learning at the beginning, for the first 25\% of the epochs we do not impose $\widehat{\text{AIB}}$ or $\widehat{\text{XIB}}$, and gradually anneal up both $\beta_1$ and $\beta_2$ during the 25\% - 50\% epochs of training, and keep them both at their final value afterwards. For the experiment in Section \ref{sec:robustness_adversary} and section \ref{sec:feature_attacks}, we perform hyperparameter search of $\beta_1\in\{0.1,0.01,0.001\}$, $\beta_2\in\{0.01,0.1\}$, $\mathcal{T}\in\{1,2\}$, $\tau\in\{0.05,0.1,1\}$, $k\in\{2,3\}$ for each dataset, and report the one with higher validation F1-micro. 
A summary of the hyperparameter scope is in Table \ref{table:hyperaparameter_GIB}. In Table \ref{table:hyperparameter_adverserial} and \ref{table:hyperparameter_adverserial_ablation}, we provide the hyperparameters that produce the results in Section \ref{sec:robustness_adversary}, and in Table \ref{table:hyperparameter_feature}, we provide hyperparameters that produce the results in Section \ref{sec:feature_attacks}.

\begin{table}[!ht]
\caption{Hyperparameter scope for Section \ref{sec:robustness_adversary} and \ref{sec:feature_attacks} for \GIBCat{} and \GIBBern{}.}
\centering
\begin{tabular}{l|c|c}
\hline
Hyperparameters      & Value/Search space & Type\\ \hline
$S_A$    & $[L]$   & Fixed$^*$       \\ 
$S_X$ & $\{L-1\}$ & Fixed     \\ 
Number $m$ of mixture components for $\mathbb{Q}(Z_X)$  & $100$ & Fixed      \\ 

$\beta_1$     & $\{0.1,0.01,0.001\}$ & Choice${}^\dagger$       \\ 
$\beta_2$      & $\{0.1,0.01\}$  & Choice          \\
$\tau$    & \{0.05,0.1,1\}    & Choice    \\ 
$k$      & $\{2,3\}$  & Choice         \\
$\mathcal{T}$      & $\{1,2\}$  & Choice         \\\hline
\multicolumn{1}{l}{$^*$Fixed: a constant value}\\
\multicolumn{1}{l}{${}^\dagger$Choice: choose from a set of discrete values}\\
\end{tabular}
\label{table:hyperaparameter_GIB}
\end{table}

\begin{table}[!ht]
\caption{Hyperparameter for adversarial attack experiment for \GIBCat{} and \GIBBern{}.}
\centering
\begin{tabular}{l|l|ccccc}
\hline
Dataset &Model       & $\beta_1$ & $\beta_2$ & $\tau$& $k$& $\mathcal{T}$ \\ \hline
\multirow{2}{*}{Cora} & \GIBCat{} & 0.001 & 0.01  &1&3&2  \\
 & \GIBBern{} & 0.001 & 0.01  &0.1&-&2  \\\hline

\multirow{2}{*}{Pubmed} & \GIBCat{} & 0.001 & 0.01  &1&3&2  \\
 & \GIBBern{} & 0.001 & 0.01  &0.1&-&2  \\\hline
 
\multirow{2}{*}{Citeseer} & \GIBCat{} & 0.001 & 0.01  &0.1&2&2  \\
 & \GIBBern{} & 0.001 & 0.01  &0.05&-&2  \\\hline
 
\end{tabular}
\label{table:hyperparameter_adverserial}
\end{table}

\begin{table}[!ht]
\caption{Hyperparameter for adversarial attack experiment for the ablations of \GIBCat{} and \GIBBern{}.}
\centering
\begin{tabular}{l|ccccc}
\hline
Model       & $\beta_1$ & $\beta_2$ & $\tau$& $k$& $\mathcal{T}$ \\ \hline
 AIB-Cat & - & 0.01  &1&3&2  \\
 AIB-Bern & - & 0.01  &0.1&-&2  \\
  XIB & 0.001 & -  &-&-&2  \\
 \hline

\end{tabular}
\label{table:hyperparameter_adverserial_ablation}
\end{table}

\begin{table}[!ht]
\caption{Hyperparameter for feature attack experiment (Section \ref{sec:feature_attacks}) for \GIBCat{} and \GIBBern{}.}
\centering
\begin{tabular}{l|l|ccccc}
\hline
Dataset &Model       & $\beta_1$ & $\beta_2$ & $\tau$& $k$& $\mathcal{T}$ \\ \hline
\multirow{2}{*}{Cora} & \GIBCat{} & 0.01 & 0.01  &0.1&2&2  \\
& AIB-Cat & - & 0.01  &0.1&2&2  \\
 & \GIBBern{} & 0.001 & 0.01  &0.05&-&2  \\
 & AIB-Bern & - & 0.01  &0.05&-&2  \\
 \hline

\multirow{2}{*}{Pubmed} & \GIBCat{} & 0.001 & 0.01  &1&3&2  \\
& AIB-Cat & - & 0.01  &1&3&2  \\
 & \GIBBern{} & 0.01 & 0.01  &0.05&-&1  \\
  & AIB-Bern & - & 0.01  &0.05&-&1  \\
 \hline
 
\multirow{2}{*}{Citeseer} & \GIBCat{} & 0.001 & 0.01  &0.1&2&2  \\
& AIB-Cat & - & 0.01  &0.1&2&2  \\
 & \GIBBern{} & 0.1 & 0.01  &0.05&-&2  \\
  & AIB-Bern & - & 0.01  &0.05&-&2  \\
 \hline
 
\end{tabular}
\label{table:hyperparameter_feature}
\end{table}

\subsection{Implementation Details for GCN and GAT}
\label{app:implementation_GCN_GAT}
We follow the default setting of GCN \cite{kipf2017semi} and GAT \cite{velickovic2018graph}, as implemented in \url{https://github.com/rusty1s/pytorch_geometric/blob/master/examples/gcn.py} and \url{https://github.com/rusty1s/pytorch_geometric/blob/master/examples/gat.py}, respectively. Importantly, we keep the dropout on the attention weights as the original GAT. Whenever possible, we keep the same architecture choice between GAT and \GIBCat{} (and \GIBBern{}) as detailed in Section \ref{app:implementation_GIB}, for a fair comparison.

\subsection{Implementation Details for RGCN and GCNJaccard}
\label{app:implementation_defense_baseline}
We used the implementation in this repository: \url{https://github.com/DSE-MSU/DeepRobust}. We perform hyperparameter tuning for both baselines for the adversarial attack experiment in Section \ref{sec:robustness_adversary}. We first tune the latent dimension, learning rate, weight decay for both models. Specifically, we search within \{16, 32, 64, 128\} for latent dimension, \{$10^{-3}$, $10^{-2}$, $10^{-1}$\} for learning rate, and \{$10^{-4}$, $5\times10^{-4}$, $10^{-3}$\} for weight decay. For GCNJaccard, we additionally fine-tune the threshold hyperparameter which is used to decide whether two neighbor nodes are still connected. We search threshold within \{0.01, 0.03, 0.05\}. For RGCN, we additionally fine-tune the $\beta_1$ within \{$10^{-4}$, $5\times10^{-4}$, $10^{-3}$\} and $\gamma$ within \{0.1, 0.3, 0.5, 0.9\}. Please find the best set of hyperparameters for both models in Table \ref{tab:citehyper}, \ref{tab:corahyper} and \ref{tab:pubhyper}.

\begin{table}[!h]
\caption{Hyperparameter of baselines used on Citeseer dataset.}
\centering
\begin{tabular}{l|c|c}
\hline
              & RGCN & GCNJaccard \\ \hline
latent dim    & 64   & 16         \\ 
learning rate & $10^{-2}$ & $10^{-2}$    \\ 
weight dacay  & $5\times 10^{-4}$ & $5\times 10^{-4}$      \\ 
threshold     & -    & $5\times 10^{-2}$       \\ 
$\beta_1$     & $5\times 10^{-4}$ & -          \\ 
$\gamma$      & $0.3$  & -          \\ \hline
\end{tabular}
\label{tab:citehyper}
\end{table}

\begin{table}[!h]
\caption{Hyperparameter of baselines used on Cora dataset.}
\centering
\begin{tabular}{l|c|c}
\hline
              & RGCN & GCNJaccard \\ \hline
latent dim    & 64   & 16         \\ 
learning rate & $10^{-2}$ & $10^{-2}$      \\ 
weight dacay  & $5\times10^{-4}$ & $5\times10^{-4}$      \\ 
threshold     & -    & $5\times10^{-2}$       \\ 
$\beta_1$     & $5\times10^{-4}$ & -          \\ 
$\gamma$      & 0.3  & -          \\ \hline
\end{tabular}
\label{tab:corahyper}
\end{table}

\begin{table}[!h]
\caption{Hyperparameter of baselines used on Pubmed dataset.}
\centering
\begin{tabular}{l|c|c}
\hline
              & RGCN & GCNJaccard \\ \hline
latent dim    & 16   & 16         \\ 
learning rate & $10^{-2}$ & $10^{-2}$       \\ 
weight dacay  & $5\times10^{-4}$ & $5\times10^{-4}$       \\ 
threshold     & -    & $5\times10^{-2}$       \\ 
$\beta_1$     & $5\times10^{-4}$ & -          \\ 
$\gamma$      & 0.1  & -          \\ \hline
\end{tabular}
\label{tab:pubhyper}
\end{table}

\subsection{Additional Details for Adversarial Attack Experiment}
\label{app:implementation_nettack}
We use the implementation of Nettack \cite{zugner2018adversarial} in the repository \url{https://github.com/DSE-MSU/DeepRobust} with default settings. As stated in the main text, for each dataset we select 40 nodes in the test set to attack with 10 having the highest margin of classification, 10 having the lowest margin of classification (but still correctly classified), and 20 random nodes. For each target node, we independently train a different model and evaluate its performance on the target node in both evasive and poisoning setting. Different from \cite{zugner2018adversarial} that only keeps the largest connected component of the graph and uses random split, to keep consistent settings across experiments, we still use the full graph and standard split, which makes the defense even harder than that in \cite{zugner2018adversarial}. For each dataset and each number of perturbations (1, 2, 3, 4), we repeat the above experiment 5 times with random seeds 0, 1, 2, 3, 4, and report the average accuracy on the targeted nodes (therefore, each cell in Table \ref{table:defense_adverserial} is the mean and std. of the performance of 200 model instances (5 seeds $\times$ 40 targeted nodes, each training one model instance). Across the 5 runs of the experiment, the 20 nodes with highest and lowest margin of classification are kept the same, and the 20 random nodes are sampled randomly and then fixed. We also make sure that for the same seed, different models are evaluated against the same 40 target nodes, to eliminate fluctuation between models due to random sampling.

\subsection{Additional Details for Feature Attack Experiment}
\label{sec:additinal_feature}
As before, for each model to compare, we train 5 instances with seeds 0, 1, 2, 3, 4. After training, for each seed and each specified feature noise ratio $\lambda$, we perform 5 random node feature attacks, by adding independent Gaussian noise $\lambda\cdot r\cdot\epsilon$ to each dimension of the node feature, where $r$ is the mean of the maximum value of each node's feature, and $\epsilon\sim N(0,1)$. Therefore, each number in Table \ref{table:feature_noise} is the mean and std. of 25 instances (5 seeds $\times$ 5 attacks per seed).

\section{Training time for \GIBCat{} and \GIBBern{}}
The training time of \GIBCat{} and \GIBBern{} is on the same order as GAT with the same underlying architecture. For example, with 2 layers, \GIBCat{} takes 98s (\GIBBern{} takes 84s) to train 2000 epochs on a NVIDIA GeForce RTX 2080 GPU, while GAT takes 51s to train on the same device. The similar order of training time is due to that they have similar number of parameters and complexity. Compared to GAT, \GIBCat{} and \GIBBern{} introduce minimal more parameters. In this work, on the structural side, we use the attention weights of GAT as parameters to encode structural representation, which keeps the same number of parameters as GAT. On the feature side, we set $S_X=\{L-1\}$, which only requires to predict the diagonal variance of the Gaussian in addition to the mean, which introduce small number of parameters. Therefore, in total, \GIBCat{} and \GIBBern{} have similar complexity. The added training time is due to the sampling of edges and node features during training. We expect that when GIB is applied to other GNNs, the augmented model has similar complexity and training time.

\section{Additional experiments for Deep Graph Infomax (DGI)}

Here we perform additional experiment for adversarial attacks on Cora using Nettack. The result is in Table \ref{table:defense_adverserial_DGI}. We see that both \GIBCat{} and \GIBBern{} outperform DGI by a large margin.

\begin{table}[t]
\caption{
Average classification accuracy (\%) for the targeted nodes under direct attack for Cora. Each number is the average accuracy for the 40 targeted nodes for 5 random initialization of the experiments. Bold font denotes top two models.
}
\label{table:defense_adverserial_DGI}
\resizebox{\textwidth}{!}{
\begin{tabular}{|l|c|cccc|cccc|}
\hline
  & \textbf{Clean (\%)}     & \multicolumn{4}{c|}{\textbf{Evasive (\%)}}                                                        & \multicolumn{4}{c|}{\textbf{Poisoning (\%)}}                                                      \\ \cline{2-10} 
                                                          &    \multicolumn{1}{l|}{}   & \textbf{1}             & \textbf{2}             & \textbf{3}             & \textbf{4}             & \textbf{1}             & \textbf{2}             & \textbf{3}             & \textbf{4}             \\ \hline
 
                                   DGI                                                  & \textbf{83.2}\scriptsize{$\pm$4.82}           & 54.5\scriptsize{$\pm$4.81}          & 41.5\scriptsize{$\pm$2.24}          & 35.5\scriptsize{$\pm$5.42}          & 31.0\scriptsize{$\pm$3.79}           & 53.5\scriptsize{$\pm$7.42}          & 38.5\scriptsize{$\pm$4.18}          & 33.0\scriptsize{$\pm$5.42}          & 29.0\scriptsize{$\pm$3.79}          \\ \cline{1-10} 
                                   \textbf{GIB-Cat}                                     & 77.6\scriptsize{$\pm$2.84} & \textbf{63.0}\scriptsize{$\pm$4.81} & \textbf{52.5}\scriptsize{$\pm$3.54} & \textbf{44.5}\scriptsize{$\pm$5.70} & \textbf{36.5}\scriptsize{$\pm$6.75} & \textbf{60.0}\scriptsize{$\pm$6.37} & \textbf{50.0}\scriptsize{$\pm$2.50}  & \textbf{39.5}\scriptsize{$\pm$5.42} & \textbf{30.0}\scriptsize{$\pm$3.95} \\
                       
                        \textbf{GIB-Bern}
                        &
                        \textbf{78.4}\scriptsize{$\pm$4.07} & \textbf{64.0}\scriptsize{$\pm$5.18} & \textbf{51.5}\scriptsize{$\pm$4.54} & \textbf{43.0}\scriptsize{$\pm$3.26} & \textbf{37.5}\scriptsize{$\pm$3.95} & \textbf{61.5}\scriptsize{$\pm$4.18} & \textbf{46.0}\scriptsize{$\pm$4.18}  & \textbf{36.5}\scriptsize{$\pm$4.18} & \textbf{31.5}\scriptsize{$\pm$2.85} \\
                                   \hline
\end{tabular}
}
\vspace{-0.1cm}
\end{table}

\section{More Detailed Analysis of Adversarial Attack in Section \ref{sec:robustness_adversary}}
\label{app:citeseer_analysis}
Table \ref{tab:analysis} summarizes the statistics of the target nodes and the adversarial perturbations by Nettack, for Cora, Pubmed and Citeseer. 

\begin{table}[!h]
\caption{Statistics of the target nodes and adversarial perturbations by Nettack in Section \ref{sec:robustness_adversary}.}
\centering
\begin{tabular}{l|ccc}
\hline
              & \textbf{Cora} & \textbf{Pubmed} & \textbf{Citeseer} \\ \hline
Fraction of degree $1$ in target nodes & 0.215 & 0.425 & \textbf{0.500}\\
Fraction of degree $\le2$ in target nodes & 0.345 & 0.565 & \textbf{0.710} \\
Fraction of degree $\le3$ in target nodes & 0.455 & 0.630 & \textbf{0.755}\\
Fraction of degree $\le4$ in target nodes & 0.540 & 0.640 & \textbf{0.810}\\\hline
Fraction of structural attacks & 1.000 & 1.000 & 0.991\\
Fraction of added-edge attack in structural attacks & 0.890 & 0.834 & 0.909 \\
Fraction of different classes in added-edge attacks & 1.000 & 0.993 & 0.985 \\
\hline
\end{tabular}
\label{tab:analysis}
\end{table}

We have the following observations:
\begin{itemize}
\item Compared to Cora and Pubmed, Citeseer has much more nodes with degrees less than 1, 2, 3, 4. This explains why in general the 5 models has worse performance in Citeseer than in Cora and Pubmed.
\item Almost all attacks ($\ge99.1\%$) are structural attacks.
\item Within structural attacks, most of them ($\ge83.4\%$) are via adding edges, with Citeseer having the largest fraction.
\item For the added edges, almost all of them ($\ge98.5\%$) have different classes for the end nodes.
\end{itemize}

From the above summary, we see that the target nodes in Citeseer dataset in general have fewest degrees, which are most prone to added-edge structural attacks by connecting nodes with different classes. This exactly satisfies the assumption of GCNJaccard \cite{wu2019adversarial}. GCNJaccard proceeds by deleting edges with low feature similarity, so those added edges are not likely to enter into the model training during poisoning attacks. This is probably the reason why in Nettack poisoning mode in Citeseer, GCNJaccard has the best performance.



\end{document}